\documentclass[runningheads]{llncs}

\usepackage{eccv}

\usepackage{eccvabbrv}

\usepackage{enumitem}
\usepackage{multirow}
\usepackage{colortbl}
\usepackage{microtype}
\usepackage{comment}
\usepackage[hang,flushmargin]{footmisc}

\usepackage{amsfonts}

\setlist[itemize]{align=parleft,left=0pt,topsep=1mm,itemsep=0mm,parsep=1mm}

\definecolor{postechred}{rgb}{0.784, 0.003, 0.313}

\definecolor{ballblue}{rgb}{0.13, 0.67, 0.8}
\definecolor{cornellred}{rgb}{0.7, 0.11, 0.11}
\definecolor{darkcyan}{rgb}{0.0, 0.55, 0.55}
\definecolor{CuGray}{gray}{0.9}
\definecolor{airforceblue}{rgb}{0.36, 0.54, 0.66}
\definecolor{rev}{rgb}{0.784, 0.003, 0.313}
\definecolor{pink}{cmyk}{0, 0.7808, 0.4429, 0.1412}
\definecolor{amethyst}{rgb}{0.6, 0.4, 0.8}
\definecolor{black}{rgb}{0.0, 0.0, 0.0}
\definecolor{dimgray}{rgb}{0.41, 0.41, 0.41}
\definecolor{bleudefrance}{rgb}{0.19, 0.55, 0.91}
\definecolor{blue(ryb)}{rgb}{0.01, 0.28, 1.0}
\definecolor{Gray}{gray}{0.88}
\definecolor{green(ncs)}{rgb}{0.0, 0.62, 0.42}
\definecolor{brightpink}{rgb}{1.0, 0.0, 0.5}
\definecolor{alizarin}{rgb}{0.82, 0.1, 0.26}
\definecolor{orange-red}{rgb}{1.0, 0.27, 0.0}
\definecolor{nicegreen}{rgb}{0.0, 0.7, 0.1}

\definecolor{kellygreen}{rgb}{0.3, 0.73, 0.09}

\newcolumntype{g}{>{\columncolor{CuGray}}c}
\newcolumntype{z}{>{\columncolor{CuGray}}l}

\renewcommand{\paragraph}[1]{\vspace{1mm}\noindent\textbf{#1.}\,}

\usepackage{xspace}

\makeatletter
\def\@fnsymbol#1{\ensuremath{\ifcase#1\or *\or \dagger\or \ddagger\or
   \mathsection\or \mathparagraph\or \|\or **\or \dagger\dagger
   \or \ddagger\ddagger \else\@ctrerr\fi}}
\makeatother

\def\onedot{.\@\xspace}
\def\eg{\emph{e.g}\onedot}

\newcommand{\Sref}[1]{Sec.~\ref{#1}}

\newcommand{\Fref}[1]{Fig.~\ref{#1}}
\newcommand{\Tref}[1]{Table~\ref{#1}}

\newcommand{\be}{\begin{eqnarray}}
\newcommand{\ee}{\end{eqnarray}}
\newcommand{\bee}{\begin{eqnarray*}}
\newcommand{\eee}{\end{eqnarray*}}

\newcommand{\matrixb}{\left[ \begin{array}}
\newcommand{\matrixe}{\end{array} \right]}

\usepackage{amssymb}
\usepackage{pifont}

\newcommand{\maketitlesupplementary}{
    \begin{center}
        {\LARGE \bf \@title\par}
        \vspace{1em}
        {\large Supplementary Material\par}
        \vspace{1em}
        {\large \@author\par}
    \end{center}
    \vspace{2em}
}

\usepackage{algorithm}
\usepackage{algpseudocode}

\usepackage{makecell}
\usepackage{wrapfig}

\newcommand{\nickname}{Ref-GeNVS\xspace}

\definecolor{gu}{rgb}{0.5460, 0.1755, 0.2766}
\definecolor{clova}{rgb}{0.24, 0.63, 0.33}

\makeatletter
\DeclareRobustCommand\onedot{\futurelet\@let@token\@onedot}
\def\@onedot{\ifx\@let@token.\else.\null\fi\xspace}
\def\eg{\emph{e.g}\onedot}

\makeatletter
\newcommand{\startsuptoc}{%
  \let\orig@addcontentsline\addcontentsline
  \def\tocfile{toc}%
  \def\addcontentsline##1##2##3{%
    \orig@addcontentsline{##1}{##2}{##3}%
    \def\temp{##1}%
    \ifx\temp\tocfile
      \orig@addcontentsline{suptoc}{##2}{##3}%
    \fi
  }%
}

\newcommand{\stopsuptoc}{%
  \let\addcontentsline\orig@addcontentsline
}

\newcommand{\tablesupcontents}{%
  \begingroup
    \renewcommand{\contentsname}{Contents (Supplementary)}
    \@starttoc{suptoc}
  \endgroup
}
\makeatother

\usepackage{graphicx}
\usepackage{booktabs}

\usepackage[accsupp]{axessibility}  

\usepackage[table]{xcolor}

\definecolor{pastelblue}{RGB}{225, 233, 250}
\definecolor{lightblue}{RGB}{235, 247, 255}

\usepackage{hyperref}

\usepackage{orcidlink}

\begin{document}

\title{Reflection-aware Generative Novel View Synthesis} 

\titlerunning{Ref-GeNVS}

\author{GeonU Kim\orcidlink{0009-0009-0224-0060} \and
Shin Dong-Yeon\orcidlink{0009-0000-1764-7829} \and
Tae-Hyun Oh\orcidlink{0000-0003-0468-1571}
}


\authorrunning{Kim et al.}

\institute{KAIST\\
\email{\{geonukim,shindy,taehyun.oh\}@kaist.ac.kr}}

\maketitle
\includegraphics[width=0.97\linewidth]{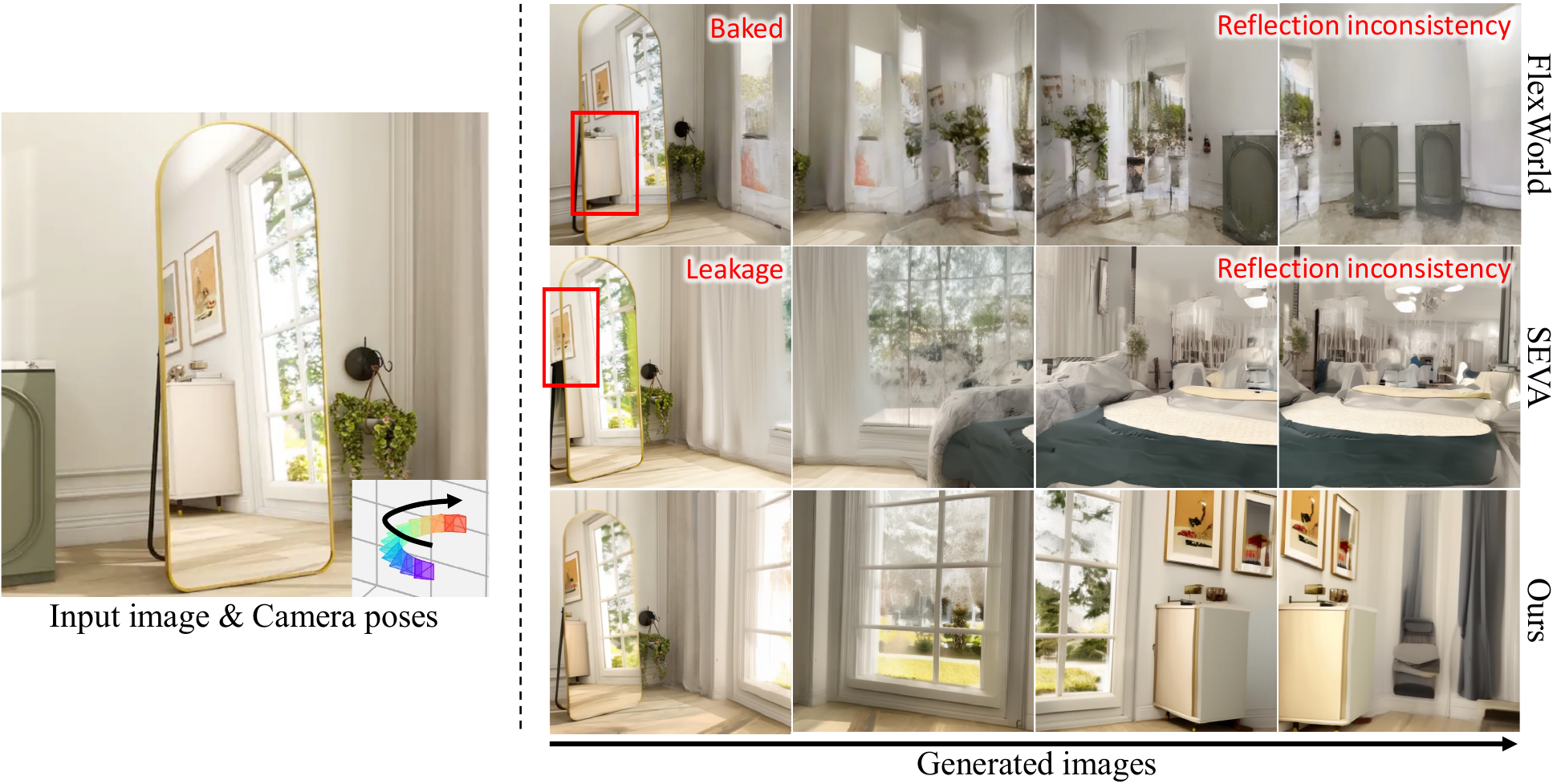}
\captionof{figure}{
    Reflection-aware generative novel view synthesis on a single image containing a mirror. Given a single image and target novel view camera poses as input (left), both FlexWorld~\cite{chen2025flexworld} and Stable Virtual Camera (SEVA)~\cite{zhou2025stable} (top two rows) fail to correctly interpret the presence of a mirror, resulting in baked mirror surfaces, leakage artifacts, or inconsistent reflections in the generated views. \nickname (bottom row) generates both the mirror surface and the scene revealed in the reflection accurately, producing photorealistic and view-consistent results from single or sparse input images.
    }
\label{fig:teaser}    
\begin{abstract}
    
We propose \nickname, a training-free, reflection-aware method for generative novel view synthesis (NVS) in mirror scenes. 
Existing multi-view diffusion models 
often fail to recognize the mirror in the scene and cannot exploit reflected content for scene generation. 
To fix this issue without additional training, our key idea is to 
treat a mirror image as two complementary views. From input images, we estimate the mirror plane and reflect camera poses to form virtual views. 
Based on this virtual view setup, we propose a two-stage generation method consisting of Mirror-gated attention and Reflection injection, which enables reflection-consistent NVS by explicitly leveraging reflection relationships in a multi-view diffusion model.
\nickname inherits the strong generalizability of the multi-view diffusion backbone, while it does not require finetuning.
On synthetic and real scenes including mirrors, \nickname outperforms recent generative NVS methods by generating reflection-consistent and contextually coherent novel views, revealing scene structure visible only through mirrors. Project page: https://kim-geonu.github.io/Ref-GeNVS/


\end{abstract}
\section{Introduction}
\label{sec:intro}

Mirrors extend humans' limited field of view, allowing  us to directly see regions that would otherwise be invisible.
More specifically, 
humans treat the mirror as an additional viewpoint by separating the mirror from the background in their view, and extrapolate the rest of the scene from the observed evidence.
This reflection-aware reasoning is critical in practice, from safety scenarios such as discovering hidden objects around road corners using mirrors, to immersive VR/AR experiences requiring faithful scene reconstruction. 
Despite this capability, we observe 
that 
state-of-the-art generative novel view synthesis (NVS) methods~\cite{chen2025flexworld, zhou2025stable} fall short of 
utilizing mirror cues and 
typically 
treat a mirror image as a single ordinary view. 
As a result, images with mirrors are handled as if the reflections provide no 
structural 
information, which leads to scene predictions that disagree with the true mirror content, baked mirror surface, and leakage from incorrect separation of reflective and non-reflective regions (see \Fref{fig:teaser}).


A straightforward approach for reflection-aware generative NVS would be to finetune large generative models on datasets containing mirrors. However, reflection-aware NVS is a physically constrained generation problem, and finetuning alone does not guarantee that models will learn the underlying reflection constraints correctly~\cite{kang2024far, dhiman2025mirrorverse, dhiman2025reflecting, hyun2026pavas}. Furthermore, collecting large-scale mirror datasets is challenging, and finetuning large models remains computationally expensive.

To circumvent these limitations, we propose \nickname, a training-free method for reflection-aware generative NVS in scenes that contain mirrors. 
\nickname uses a multi-view diffusion backbone \cite{zhou2025stable, cao2025mvgenmaster} and is designed to incorporate reflected evidence even when only mirror-containing inputs are available.
To explicitly inject reflected evidence into the backbone, we first estimate the plane equation of the mirror in the scene, and reflect camera poses of input views to the mirror to construct virtual reflected views.
Then we generate the target views via a two-stage generation pipeline:
(1) generate target views with mirror-masked input images and reflected virtual views while restricting conditioning to the mirror regions of the reflected views using \textit{Mirror-gated attention}; and (2) complete the mirror surface via \textit{Reflection injection} by injecting features from the reflected target poses into the mirror region at each denoising step. 
This approach closes the gap between human reflection-aware reasoning and current NVS models, enabling photorealistic and reflection-consistent novel views without finetuning.

For evaluation, we construct a dataset of realistic 3D scenes with mirrors to probe reflection handling, and we also test on a real scene dataset~\cite{zeng2023mirror-nerf}. Across both 
data, \nickname\ generates photorealistic, reflection-consistent novel views from single and sparse inputs, surpassing recent generative NVS methods.

\begin{itemize}
    \item We propose 
    a training-free pipeline for 
    \textit{the first} reflection-aware generative NVS method, by treating a mirror image as two complementary views to generate the scene reflected on the mirror. 
    \item We introduce Mirror-gated attention and Reflection injection, two techniques designed to 
    explicitly
    leverage reflected cues during generation and improve reflection-consistent novel-view synthesis without model finetuning.
\end{itemize}

\section{Related Work}
\label{sec:related_work}


\paragraph{Reflection-aware novel view synthesis} Novel view synthesis (NVS) renders photorealistic images from unseen viewpoints given calibrated inputs, typically via implicit radiance fields or explicit point-based primitives for volume rendering, with Neural Radiance Fields (NeRF) and 3D Gaussian Splatting (3DGS) as representative baselines \cite{mildenhall2020nerf, kerbl20233DGS, barron2021mip ,barron2022mipnerf360,Chen2022TensoRF, mueller2022instantngp, yu2022plenoxels, li2024dngaussian, zhu2023fsgs, jun2022hdr, dong2026hdr}.
Although these methods achieve impressive reconstruction quality, they often misinterpret reflective objects since strong view-dependent appearances violate the multi-view photometric consistency assumed by NeRF- and 3DGS-based pipelines. As a result, reflections are treated as outliers and the geometric correspondence induced by mirrors is not exploited.
To overcome this limitation, prior reflection modeling separates diffuse and specular components with physically based decomposition and environment lighting estimation \cite{liu2023nero,jiang2023gaussianshader}, or renders real and reflected content using two radiance fields or reparameterized heads and blends the outputs \cite{verbin2022refnerf,Guo2022NeRFReN, wang2023unisdf}. 
Mirror-NeRF \cite{zeng2023mirror-nerf}, MS-NeRF \cite{Yin2025msnerf}, TraM-NeRF~\cite{van2023tram}, and MirrorGaussian \cite{liu2024mirrorgaussian} reconstruct mirrors by reflecting camera rays about the surface normal of the mirrors and rendering them, using volumetric ray marching for NeRF variants and point-based rasterization for Gaussian splatting.
These methods only focus on rendering reflective objects themselves, whereas our main goal is to synthesize the remainder of the scene using a mirror and generate the surface of the mirror at the same time.

Our setting is therefore also related to OrCa~\cite{tiwary2023orca} and World-from-Eyes~\cite{ alzayer2024seeing}, which can be framed as NVS applied to passive non-line-of-sight imaging. 
These methods use a multi-view captured glossy object as a proxy mirror to render the surrounding environment. 
\nickname departs from these methods in two key aspects. Our method operates on extremely sparse input images (one or three input images) and assumes an ideal planar mirror.

\paragraph{Generative novel view synthesis}
Generative novel view synthesis leverages strong priors from 2D/3D diffusion models to synthesize multi-view-consistent, photorealistic images from single or sparse input views~\cite{chan2023genvs, liu2023zero123, zeronvs, gao2024cat3d, wu2023reconfusion, zhou2025stable,sun2024dimensionx, Shi2024MVDream}.
Another line of work adopts a warp-and-inpaint paradigm: they estimate coarse geometry, warp the input view into the target view, and then apply diffusion to inpaint holes from occlusions and out-of-frame regions~\cite{yu2024viewcrafter, cao2025mvgenmaster,  you2024solver, chung2023luciddreamer, zhang2024text2nerf, seo2024genwarp,ma2025you, chen2025flexworld, ren2025gen3c, wang2025vistadream}.
However, none of these methods are trained to handle mirrors. As a result, reflective regions are often misinterpreted as background, producing views inconsistent with the correct mirror reflections.
In contrast, \nickname explicitly incorporates mirror information into a multi-view diffusion backbone, enabling reflection-consistent novel view synthesis without additional finetuning.

\begin{figure*}[!t]
    \centering
    \includegraphics[width=0.9\linewidth]{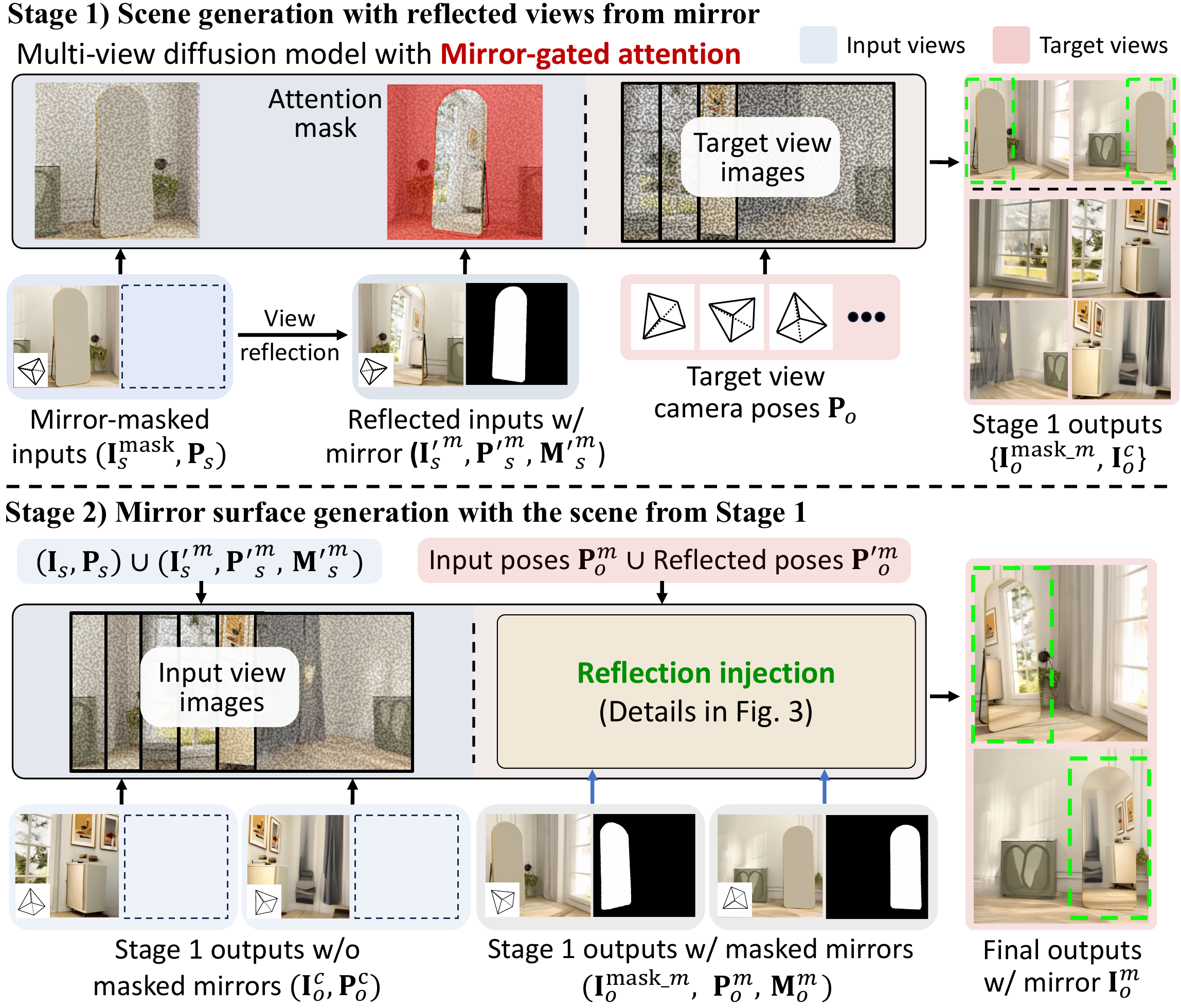}
    \caption{
    Overview of \nickname. In Stage 1, we first mask mirror surface to prevent interference between real-scene and reflected geometry. Then we generate the reflected scene by mirror-masked inputs and the reflected virtual views which consist of flipped input images, reflected camera poses, and the mirror masks. We propose \emph{Mirror-gated attention} enabling only pixels inside the mirror influence the reflected views.  In Stage 2, we perform \emph{Reflection injection} process which generates the mirror surface using the images generated in Stage 1 as inputs. 
    Finally, we obtain complete target images $\textbf{I}_o$ from target views, which are consistent with the scene reflected on the mirror. 
    }
    \label{fig:overview}
\end{figure*}

\section{Method}
\label{sec:method}


Given input images $\textbf{I}_s$, our goal is twofold: to generate target views $\textbf{I}_o$ which are consistent with the scene content revealed in the mirror, and to generate the mirror surface in $\textbf{I}_o$ itself being globally consistent with the generated scene.
The core idea of our method is providing additional information from the reflected region of the input images to the multi-view diffusion model, by obtaining the camera poses of reflected views.


In preprocess, we obtain reflected virtual views by reflecting the camera poses of input views containing mirrors across the estimated mirror plane. We then perform a two-stage generation process, as shown in \Fref{fig:overview}. Stage 1 synthesizes the target scene from mirror-masked input views and reflected virtual views~(\Sref{subsec:mirror-gated attention}), while Stage 2 restores the mirror appearance to be consistent with the generated scene~(\Sref{subsec:stage3}). We also introduce an automatic mirror plane estimation pipeline for preprocessing (Sec.~\ref{sec:automation}).

\paragraph{Preprocess}
Given input images $\textbf{I}_s$ with binary mirror masks $\textbf{M}_s$, camera poses $\textbf{P}_s$ and a mirror plane $\mathbf{e}=[\mathbf{n},d]$, we produce reflected view tuples $(\textbf{I}'^m_s,\textbf{P}'^m_s,\textbf{M}'^m_s)$ for later stages.
With the mirror plane $\mathbf{e}$, each input pose from view including mirror $\textbf{P}^m_{s}$ is reflected across the plane using Householder reflection~\cite{householder2006theory} to obtain the pose of a virtual camera in the mirror. Then we apply horizontal flip to the both image/mask pair and the corresponding reflected camera pose for maintaining the same-hand coordinate convention as the original inputs. 
As a result, we obtain reflected views with the mirror masks $(\textbf{I}'^m_s, \textbf{P}'^m_s,\ \textbf{M}'^m_s)$, where $(\textbf{I}'^m_s, \textbf{P}'^m_s)$ are additional inputs for multi-view diffusion model and the masks $\textbf{M}'^m_s$ are used in \textit{Mirror-gated attention} process.


\subsection{Preliminary: Multi-view diffusion model}
\label{subsec:preliminary}
We adopt an ``$N$-in $M$-out'' multi-view diffusion model as our backbone, which aims to generate novel views sampled from:
\begin{equation}
\label{eq:SEVA}
p\!\left(\mathbf{I}_{o}\,\middle|\,(\mathbf{I}_{s}, \mathbf{P}_{s}),\, \mathbf{P}_{o}\right),
\end{equation}
where $\mathbf{I}_{s}, \mathbf{I}_{o}$ denote the $N$ input and $M$ target images and $\mathbf{P}_{s}, \mathbf{P}_{o}$ are their camera poses, respectively. 
Multi-view diffusion models~\cite{gao2024cat3d,cao2025mvgenmaster,zhou2025stable} 
use cross-view attention,
which connects all different views in self attention by inflating the original 2D self-attention. 
During denoising process, the cross-view attention modules allow input and target frames to exchange information at every step, while an input/target indicator distinguishes encoded inputs from noisy targets. 

Building on the multi-view diffusion model, we develop a training-free reflection-aware NVS pipeline which introduces mirror-masked conditioning with \textit{Mirror-gated attention} and novel \textit{Reflection injection} algorithm.
Our method can be generally incorporated into multi-view diffusion models using cross-view attention~\cite{zhou2025stable, cao2025mvgenmaster} without finetuning.

\subsection{Stage 1: Scene generation 
with Mirror-gated attention
}
\label{subsec:mirror-gated attention}
Our goal is to generate target images $\textbf{I}_o$ at target poses $\textbf{P}_o$ using the reflected views $(\textbf{I}'^m_s,\textbf{P}'^m_s)$ obtained from the preprocess as additional inputs, which can be denoted as:
\begin{equation}
\label{eq:SEVA}
p_{\text{\nickname}}\!\left(\textbf{I}_o \,\middle|\, (\textbf{I}_s,\textbf{P}_s),\, (\textbf{I}'^m_s,\textbf{P}'^m_s),\, \textbf{P}_o\right).
\end{equation} The concept is simple, however, naively feeding reflected views as additional views to a multi-view diffusion model presents three issues: \begin{enumerate}
\item \textit{Mirror recognition.} The model may not recognize the mirror separate it from the non-reflective region, leading to baked surfaces or leakage (see \Fref{fig:teaser}).
\item \textit{Incorrect conditioning from non-mirror pixels.} If reflected images are used as inputs, pixels outside the mirror may also act as reflected evidence and incorrectly influence target-view generation.
\item \textit{Reflection inconsistency in the generated mirror surface.} Since the mirror surface and the rest of the scene are generated independently, the synthesized target view can become inconsistent with the scene reflected in the mirror.

\end{enumerate}
To address these issues, Stage 1 first masks mirror regions with a uniform color to explicitly separate mirrors from the background. The masked input views and reflected virtual views are then processed with \textit{Mirror-gated attention}, which restricts attention to mirror-region tokens in the reflected inputs. In Stage 2, we perform \textit{Reflection injection}, a guided inpainting process that uses the scene synthesized in Stage 1 to complete the mirror surface, ensuring that the synthesized reflections remain consistent with the generated scene.

\paragraph{Problem setup of Stage 1} As shown in \Fref{fig:overview} Stage~1,
we use two input sets: (i) \emph{mirror-masked original inputs} $(\textbf{I}^{\mathrm{mask}}_s,\textbf{P}_s)$ formed by masking the mirror region $\textbf{M}^m_{s}$ in $\textbf{I}_{s}$, and (ii) the \emph{reflected views with mirror mask} $(\textbf{I}'^m_s,\textbf{P}'^m_s,\textbf{M}'^m_s)$ from the preprocess.
The multi-view diffusion model receives $\{(\textbf{I}^{\mathrm{mask}}_{s},\textbf{P}_{s})\}\cup\{(\textbf{I}'^m_{s},\textbf{P}'^m_{s},\textbf{M}'^m_{s})\}$ with $\textbf{P}_o$ and generates $\textbf{I}^{\text{mask}}_o$, which are target view images with masked mirror.


\paragraph{Mirror-gated attention for the reflected inputs}
For reflected inputs with the binary mirror masks $(\textbf{I}'^m_{s},\textbf{P}'^m_{s},\textbf{M}'^m_{s})$, we use $\textbf{M}'^m_{s}$ to restrict attention to tokens inside the mirror in the reflected views. 
Tokens outside the mirror region are masked during the cross-view attention process, ensuring that non-reflected content does not influence target-view generation.

Formally, let $Q \in \mathbb{R}^{N_q \times d}$ denote the query tokens of the target view and $K,V \in \mathbb{R}^{N_k \times d}$ denote the key and value tokens extracted from the reflected input views. 
Standard cross-view attention is computed as:
\begin{equation}
\mathrm{Attn}(Q,K,V)=\mathrm{softmax}\!\left(\frac{QK^\top}{\sqrt{d}}\right)V .
\end{equation}
To prevent tokens outside the mirror from contributing to the attention, we construct a binary mask $m \in \{0,1\}^{N_k}$ derived from the reflected mirror mask $\textbf{M}'^m_s$. The mask is obtained by downsampling and reshaping $\textbf{M}'^m_s$ to align with the token layout of the attention module.
The attention logits are then gated by:
\begin{equation}
S'_{ij} =
\begin{cases}
\frac{q_i^\top k_j}{\sqrt{d}}, & m_j = 1 \\
-\infty, & m_j = 0 ,
\end{cases}, \ \ \ \ 
\mathrm{Attn}_{\text{mirror}}(Q,K,V)
=
\mathrm{softmax}(S')V .
\end{equation}
This masking sets the attention weight of tokens outside the mirror region to zero after the softmax, ensuring that only reflected content contributes to the target-view synthesis.
This Mirror-gated attention is applied consistently across all attention operators in the backbone, including the 3D attention and 1D attention modules~\cite{zhou2025stable} according to the model architecture.
Restricting attention prevents artifacts caused by reflecting the non-mirror region of the input images.


\begin{figure*}[t!]
    \centering
    \includegraphics[width=1.0\linewidth]{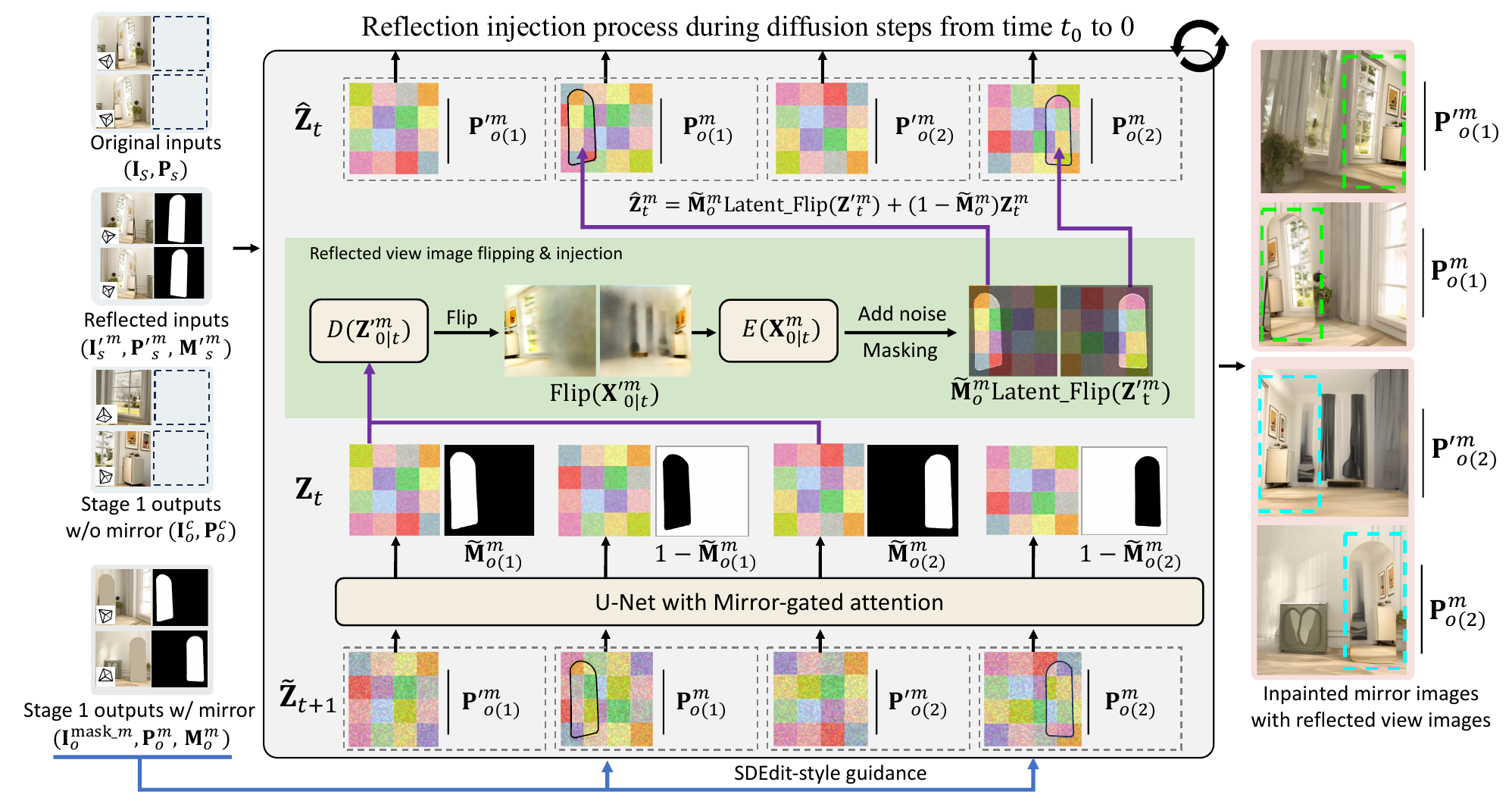}
    \caption{    
    Reflection injection process in Stage~2. We fill the mirror surface in the target views using three sources: the original inputs $(\textbf{I}_s,\textbf{P}_s)$, the reflected inputs from the preprocess $(\textbf{I}'^{\,m}_s,\textbf{P}'^{\,m}_s,\textbf{M}'^{\,m}_s)$, and the Stage~1 outputs $(\textbf{I}_o^\text{mask},\textbf{P}_o)$. 
    To generate the mirror surface in the Stage~1 outputs $(\textbf{I}_o^{\mathrm{mask}\_m},\textbf{P}_o^{m})$ consistent with the scene in reflection, we condition the model on two type of target-frame poses: the identity poses $\textbf{P}_o^{m}$ and the corresponding reflected poses $\textbf{P}'^{\,m}_o$. 
    During each denoising step $t\!\to\!t\!-\!1$, the latent feature maps $\textbf{Z}'^{\,m}_t$ predicted under $\textbf{P}'^{\,m}_o$ are horizontally flipped and injected inside the target mirror mask $\tilde{\textbf{M}}_o^{m}$ of the current target latent $\textbf{Z}^{\,m}_t$. 
    This latent injection is integrated with SDEdit-style guidance~\cite{meng2022sdedit}, enabling natural mirror-surface completion.
    The result is the inpainted mirror which is consistent with the scene generated in Stage~1 and with the reflected evidence from the original input images.
}
    \label{fig:stage3_detail}
\end{figure*}

\subsection{Stage 2: Mirror surface generation with Reflection injection}
\label{subsec:stage3}

In Stage~2, we complete the mirror surface in the outputs from Stage~1.
To complete the mirror surface with reflection-consistent content, we introduce \textit{Reflection injection} method that injects reflected-view evidence into the mirror region during the denoising process.
To further stabilize generation near the mirror boundary, we additionally employ an SDEdit-style guidance~\cite{meng2022sdedit} which enables natural compositing of mirror and non-mirror region.

\paragraph{Input setup with the outputs from Stage~1}
To inpaint the masked mirror in the Stage~1 outputs $\mathbf{I}_o^{\text{mask}}$, we first split $\mathbf{I}_o^{\text{mask}}$ into images containing mirrors and images without mirrors.
Mirror masks $\mathbf{M}_o$ are obtained using SAM2~\cite{ravi2024sam}, guided by the source-view mirror masks $(\mathbf{I}_s^{\text{mask}},\mathbf{M}_s^{m})$.
Since the mirror is already masked in Stage~1, SAM2 robustly tracks the mirror region in $\mathbf{I}_o^{\text{mask}}$.
We therefore obtain $(\mathbf{I}_o^{\text{mask}},\,\mathbf{M}_o)
=
\{(\mathbf{I}_o^{\text{mask}\_m},\,\mathbf{M}_o^{m}),
(\mathbf{I}_o^{c},\,\mathbf{M}_o^{c})\},$
where $\mathbf{I}_o^{\text{mask}\_m}$ contains mirrors and $\mathbf{I}_o^{c}$ contains no mirror regions.

\paragraph{Problem setup of Stage~2}
As illustrated in \Fref{fig:stage3_detail}, we use four types of inputs:
the original views $(\mathbf{I}_s,\mathbf{P}_s)$,
the reflected inputs $(\mathbf{I}'^{\,m}_s,\mathbf{P}'^{\,m}_s,\mathbf{M}'^{\,m}_s)$,
the Stage~1 outputs with mirrors $(\mathbf{I}_o^{\text{mask}\_m},\mathbf{P}_o^{m},\mathbf{M}_o^{m})$,
and Stage~1 outputs without mirrors $(\mathbf{I}_o^{c},\mathbf{M}_o^{c})$.
Our goal is to generate $\mathbf{I}_o^{m}$ at pose $\mathbf{P}_o^{m}$ such that the mirror region shows the reflection implied by the reflected virtual target pose $\mathbf{P}_o'^{m}$.

\begin{figure}[!t]
    \centering
    \includegraphics[width=1.0\linewidth]{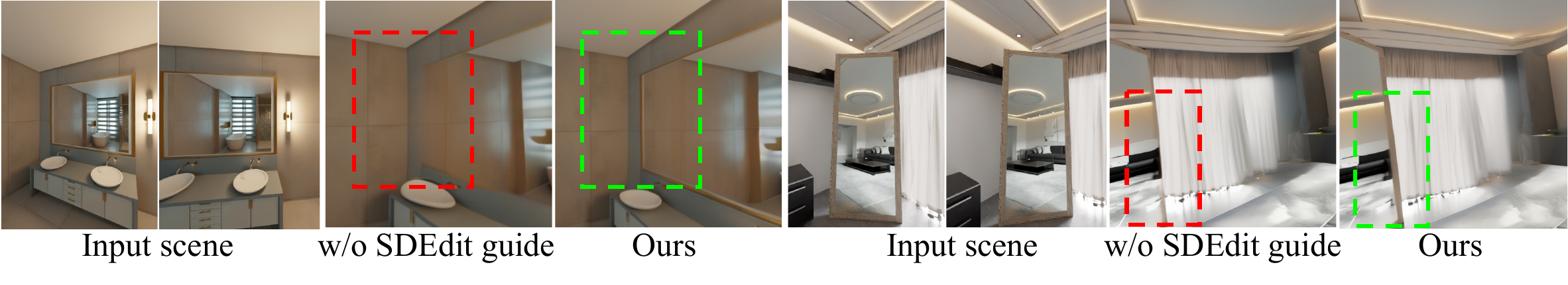}
    \caption{Effect of SDEdit-style~\cite{meng2022sdedit} guidance. Without SDEdit-style guidance, the model can produce noticeable artifacts near the boundaries of the mirror. The guidance enable natural compositing of mirror and non-mirror region.}
    \label{fig:sdedit}
\end{figure}

\paragraph{Reflection injection}
Let $\mathbf{Z}^{m}_t$ be the latent at diffusion step $t$ under pose $\mathbf{P}_o^{m}$ and $\mathbf{Z}'^{m}_t$ the latent under the reflected pose $\mathbf{P}'^{m}_o$.
We inject reflected evidence into the mirror region using the binary mask $\mathbf{M}_o^{m}$:
\begin{equation}
\label{eq:latent-injection}
\hat{\mathbf{Z}}^{m}_t
=
\tilde{\mathbf{M}}_o^{m}
\,\mathrm{Latent\_Flip}\!\left(\mathbf{Z}'^{m}_t\right)
+
(\mathbf{1}-\tilde{\mathbf{M}}_o^{m})
\,\mathbf{Z}^{m}_t ,
\end{equation}
where $\tilde{\mathbf{M}}_o^{m}$ denotes the mirror mask resized to the latent resolution.
The operator $\mathrm{Latent\_Flip}(\cdot)$ aligns the reflected latent to the target frame via decode–flip–encode:
\begin{equation}
\label{eq:latent-flip}
\mathrm{Latent\_Flip}(\mathbf{Z}'^{m}_t)
=
E\!\left(
\mathrm{Flip}\!\left(
D(\mathbf{Z}'^{m}_{0|t})
\right)
\right)
+
\sigma_t\boldsymbol{\epsilon}_t ,
\end{equation}

where $D(\cdot)$ and $E(\cdot)$ denote the VAE decoder and encoder, $\mathbf{Z}'^{m}_{0|t}$ is the predicted clean latent from $\mathbf{Z}'^{m}_t$ using Tweedie’s formula~\cite{efron2011tweedie}, and $\sigma_t$ is the noise scale at step $t$.
The decode–flip–encode route is necessary since the latent feature space is not inherently flip-equivariant with respect to the image space.

\paragraph{SDEdit-style guidance for mirror boundary stabilization}
Although reflection injection enforces reflection consistency inside the mirror, directly injecting reflected latent into the mirror region may lead to unstable generation near the mirror boundary due to abrupt transitions between injected latent and surrounding regions, shown in Fig.~\ref{fig:sdedit}.
We therefore propose an SDEdit-style guidance that anchors the non-mirror region, which enables robust mirror boundary generation.

SDEdit generates images by guiding the diffusion process from an intermediate noisy state and has been applied to image editing and composition tasks (see \cite{meng2022sdedit} for further details). Inspired by this, we extend the idea to a multi-view diffusion setting to facilitate natural composition between mirror and non-mirror regions.

We first encode the Stage~1 output $\mathbf{I}_o^{\text{mask}\_m}$ and apply forward diffusion to obtain, 
$\mathbf{Z}^{\text{init}}_t
=
\sqrt{\bar{\alpha}_t}E(\mathbf{I}_o^{\text{mask}\_m})
+
\sqrt{1-\bar{\alpha}_t}\boldsymbol{\epsilon},$ where $\bar{\alpha}_t$ denotes the cumulative noise schedule and $\boldsymbol{\epsilon}\sim\mathcal{N}(0,I)$.
During denoising, the non-mirror region is constrained by:
\begin{equation}
\label{eq:sdedit-clamp}
\tilde{\mathbf{Z}}^{m}_t
=
(\mathbf{1}-\tilde{\mathbf{M}}_o^{m})\mathbf{Z}^{\text{init}}_t
+
\tilde{\mathbf{M}}_o^{m}\hat{\mathbf{{Z}}}^{m}_t .
\end{equation}

This anchors the surrounding context while allowing the mirror region to be synthesized during generation. We apply SDEdit-style guidance only until the timestep $t_0$, to preserve the surrounding scene in the early denoising stage. From the timestep $t_0$, we instead apply reflection injection, which progressively synthesizes the mirror appearance under the guidance of the reflected inputs. The diffusion prior then naturally harmonizes the boundary between the synthesized mirror and the surrounding scene.


\subsection{Automatic preprocess}
\label{sec:automation}
\begin{figure}[!t]
    \centering
    \includegraphics[width=1.0\linewidth]{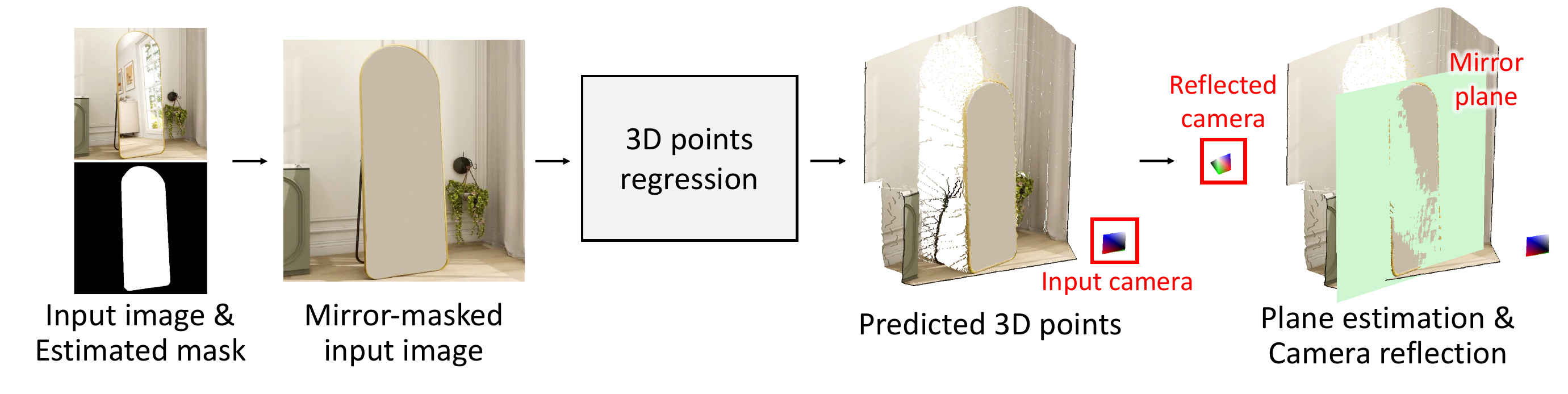}
    \caption{Automatic preprocess. We first segment mirror from the input images using DAM~\cite{xing2023DAM} and mask mirror with a uniform color. Then we estimate the mirror plane from input images by first regressing per-pixel 3D points~\cite{lin2025depthanything3} and obtaining the plane equation of the mirror using RANSAC~\cite{fischler1981random} algorithm. 
    We then reflect the input camera poses across the plane to obtain the reflected camera poses. }
    \label{fig:preprocess}
\end{figure}

We present an automated pipeline for mirror detection and mirror plane estimation in our preprocess, as shown in Fig.~\ref{fig:preprocess}. 
Since mirror detection remains a challenging problem, 
previous mirror-related NVS methods (\eg, Mirror-NeRF~\cite{zeng2023mirror-nerf}, MirrorGaussian~\cite{liu2024mirrorgaussian}) assume given mirror masks.
To relax this strong assumption while maintaining robustness to errors in the estimated mirror mask, we propose a mask-tolerant mirror plane estimation pipeline.

We first estimate binary mirror masks $\textbf{M}_s$ from input images $\textbf{I}_s$ using DAM~\cite{xing2023DAM}, an off-the-shelf mirror segmentation method.
Instead of directly exploiting mirror edge points for mirror plane estimation as in prior work~\cite{liu2024mirrorgaussian}, we mask the mirror region $\textbf{M}_s$ in $\textbf{I}_s$ with a single color and predict per-pixel 3D points $\textbf{X}_s$ using a 3D point regression method~\cite{lin2025depthanything3}. 
From the predicted per-pixel 3D points, we extract 3D points on the mirror surface $\textbf{X}_s^{\text{mirror}}$ and estimate the mirror plane equation $\mathbf{e}=[\mathbf{n},d]$ by fitting a plane using RANSAC~\cite{fischler1981random}. 
See the supplementary material for additional details and a robustness analysis of the automation process.


\begin{figure*}[!t]
    \centering
        \includegraphics[width=1\linewidth]{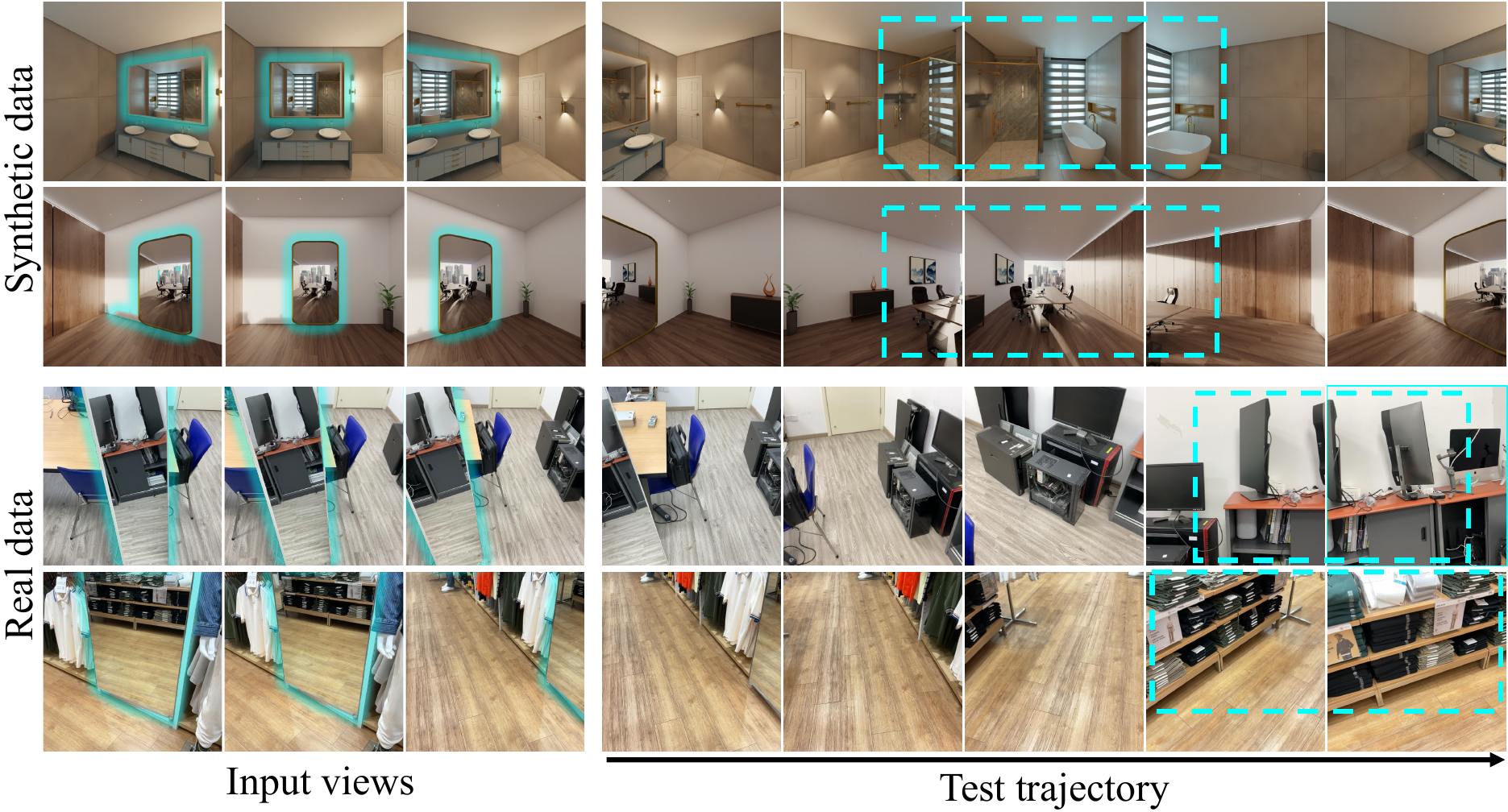}
    \caption{
    Input views and test trajectories. To evaluate reflection-aware novel view synthesis, we sample challenging, scene-specific trajectories for our synthetic dataset (top) and the real dataset~\cite{zeng2023mirror-nerf} (bottom). Each input view contains a mirror, and the trajectory includes regions visible only via reflection (red dashed box). For clarity, we overlay neon highlights to mark the mirror in the input views.
    }
    
    \label{fig:supp_dataset_trajectory}
\end{figure*}
\section{Experiments}
\label{sec:exp}
\paragraph{Dataset}
Evaluation of reflection-aware novel view synthesis (NVS) requires ground truth novel views of the scene on the opposite side of the mirror that are visible only via reflection. 
However, most of existing datasets containing mirror~\cite{dhiman2025mirrorverse, dhiman2025reflecting, wu2025reflect3r, yang2019my, warren2024effective} provide views only facing the mirror, not suitable for evaluation.
To evaluate reflection-aware NVS, we gather indoor scenes from Blender Demo~\cite{blender_demo_files}, BlenderKit~\cite{blenderkit_assets}, and CGTrader~\cite{cgtrader_assets}, and place mirrors in the scene by focusing on environments where mirrors naturally occur, such as bathrooms, and living rooms.
We also adopt the Blender scenes from the Reflect3r~\cite{wu2025reflect3r} dataset and redefine the camera viewpoints.
The scenes are rendered with physically based mirror materials, allowing realistic reflections.
We provide 8 synthetic scenes consist of ground truth multi-view images with the corresponding camera intrinsic and extrinsic parameters for evaluation. 
We also conduct experiments on the Mirror-NeRF dataset~\cite{zeng2023mirror-nerf}, which is captured in real-world.


\paragraph{Input \& test views selection}
To evaluate reflection-aware generative NVS, we design challenging input–test splits for both the synthetic dataset and the real dataset~\cite{zeng2023mirror-nerf}. As shown in \Fref{fig:supp_dataset_trajectory}, the input images include a visible mirror, and the test trajectory is chosen to contain the scene reflected on the mirror.  
For the synthetic dataset, we render each scene along an orbit trajectory showing both the mirror and the reflected scene. For the real dataset, we resample camera poses from the views provided by the original Mirror-NeRF dataset to set target views which reveal mirror-visible content.

\paragraph{Competing methods}
We evaluate our method by incorporating it into representative generative NVS approaches under sparse-input and single image NVS settings.
For sparse-input novel view synthesis, we include 
MVGenMaster~\cite{cao2025mvgenmaster}
and SEVA~\cite{zhou2025stable} and we use them as baselines of \nickname.
For single-image novel view synthesis, we similarly compare against VistaDream~\cite{wang2025vistadream}, FlexWorld~\cite{chen2025flexworld}, 
and SEVA~\cite{zhou2025stable}, and we use SEVA as our baseline module.


\subsection{Sparse-input novel view synthesis}
\begin{table}[t]
\centering
\scriptsize
\caption{Quantitative results of sparse-image novel view synthesis using DreamSim (DS)~\cite{fu2023dreamsim}, CLIP~\cite{radford2021clip} similarity score, PSNR, SSIM~\cite{ssim}, and LPIPS~\cite{zhang2018lpips}. Our method achieves better performance compared to MVGenMaster~\cite{cao2025mvgenmaster} and SEVA~\cite{zhou2025stable}.}
{\renewcommand{\arraystretch}{1.1}
\begin{tabular}{ccccccccccc}
\hline
\multicolumn{1}{l}{} & \multicolumn{5}{c|}{Real data~\cite{zeng2023mirror-nerf}} & \multicolumn{5}{c}{Synthetic data} \\ \hline
Metric               
& DS$\downarrow$ 
& CLIP$\uparrow$ 
& PSNR$\uparrow$ 
& SSIM$\uparrow$ 
& \multicolumn{1}{c|}{LPIPS$\downarrow$} 
& DS$\downarrow$ 
& CLIP$\uparrow$ 
& PSNR$\uparrow$ 
& SSIM$\uparrow$ 
& LPIPS$\downarrow$ \\ 
\hline

MVGenMaster~\cite{cao2025mvgenmaster}         
& 0.361 & 0.867 & 12.75 & 0.413 & \multicolumn{1}{c|}{0.536}
& 0.227 & 0.926 & 16.32 & 0.719 & 0.378 \\

Ours (MVGen.)         
& \cellcolor{pastelblue}0.194 & \cellcolor{pastelblue}0.930 & \cellcolor{pastelblue}13.67 & \cellcolor{pastelblue}0.455 & \multicolumn{1}{c|}{\cellcolor{pastelblue}0.474}
& \cellcolor{pastelblue}0.116 &\cellcolor{pastelblue}0.959 &\cellcolor{pastelblue}17.50 &\cellcolor{pastelblue}0.732 &\cellcolor{pastelblue}0.313 \\ 
\hline

SEVA~\cite{zhou2025stable}                 
& 0.275 & 0.920 & 12.03 & 0.394 & \multicolumn{1}{c|}{0.564}
& 0.270 & 0.924 & 13.62 & 0.612 & 0.492 \\

Ours (SEVA)                 
& \cellcolor{pastelblue}0.129 & \cellcolor{pastelblue}0.951 & \cellcolor{pastelblue}14.19 & \cellcolor{pastelblue}0.455 & \multicolumn{1}{c|}{\cellcolor{pastelblue}0.466}
& \cellcolor{pastelblue}0.156 &\cellcolor{pastelblue}0.948 &\cellcolor{pastelblue}15.27 &\cellcolor{pastelblue}0.676 &\cellcolor{pastelblue}0.403 \\

\hline
\end{tabular}
}
\label{tab:sparse-quant-table}
\end{table}

\begin{figure*}[!t]
    \centering
        \includegraphics[width=1.0\linewidth]{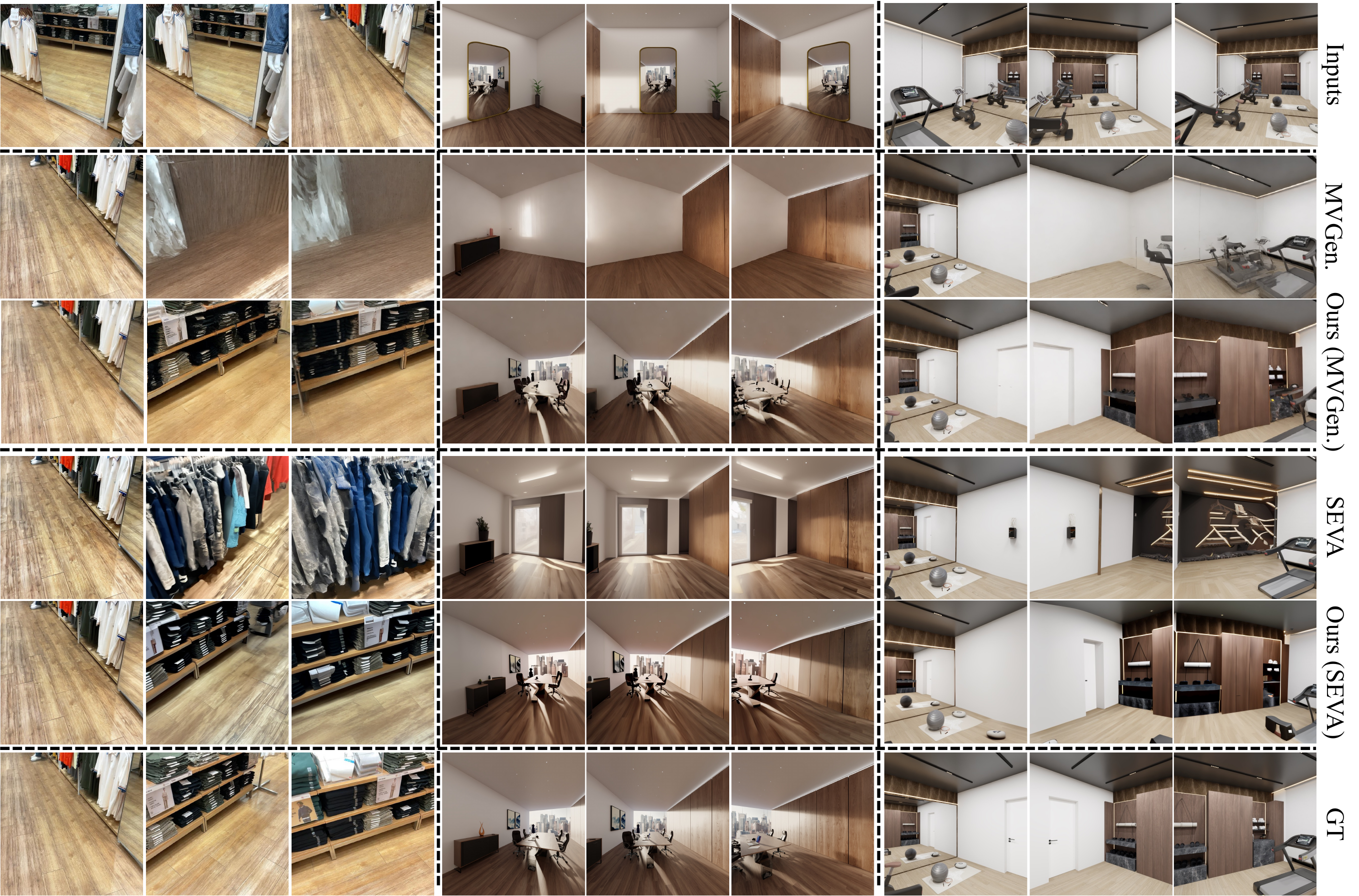}
    \caption{Qualitative comparison under sparse-input novel view synthesis. \nickname produces more consistent and reflection-aware target views compared to prior methods~\cite{zhou2025stable, cao2025mvgenmaster}, especially in regions only visible through the mirror.}
    \label{fig:sparse_view_qual}
\end{figure*}
\paragraph{Evaluation protocol in sparse-input setting} 
We evaluate competing methods using both perceptual and pixel-level metrics. We report DreamSim~\cite{fu2023dreamsim}, CLIP~\cite{radford2021clip} similarity, PSNR, SSIM~\cite{ssim}, and LPIPS~\cite{zhang2018lpips} computed between generated images and ground-truth target views.
DreamSim and CLIP similarity quantify high-level perceptual similarity while PSNR, SSIM, and LPIPS qualify geometric consistency in the scenes containing mirror.

\paragraph{Quantitative results}
As shown in \Tref{tab:sparse-quant-table},
across all quantitative metrics including DreamSim, CLIP similarity, PSNR, SSIM, and LPIPS, \nickname surpasses both MVGenMaster and SEVA by substantial margins. 
These results show that existing 
generative novel view synthesis methods are fundamentally limited when the scene requires reasoning about reflected content, while \nickname explicitly handles reflected evidence.


\paragraph{Qualitative results}
The qualitative results make this gap even more apparent. As shown in \Fref{fig:sparse_view_qual}, competing methods consistently misinterpret or ignore mirror-reflected regions, leading to severe structural inconsistencies and hallucinated content when synthesizing the target view. 
In contrast, \nickname based on the competing methods effectively leverages the reflection and produces novel views that align closely with the ground truth images. 
The fidelity and coherence of mirror-revealed regions in our outputs highlight \nickname’s advantage in reflection-dependent scenarios and its generalizability across different model architectures.

\subsection{Novel view synthesis from a single image}
\begin{figure}[!t]
    \centering
        \includegraphics[width=1.0\linewidth]{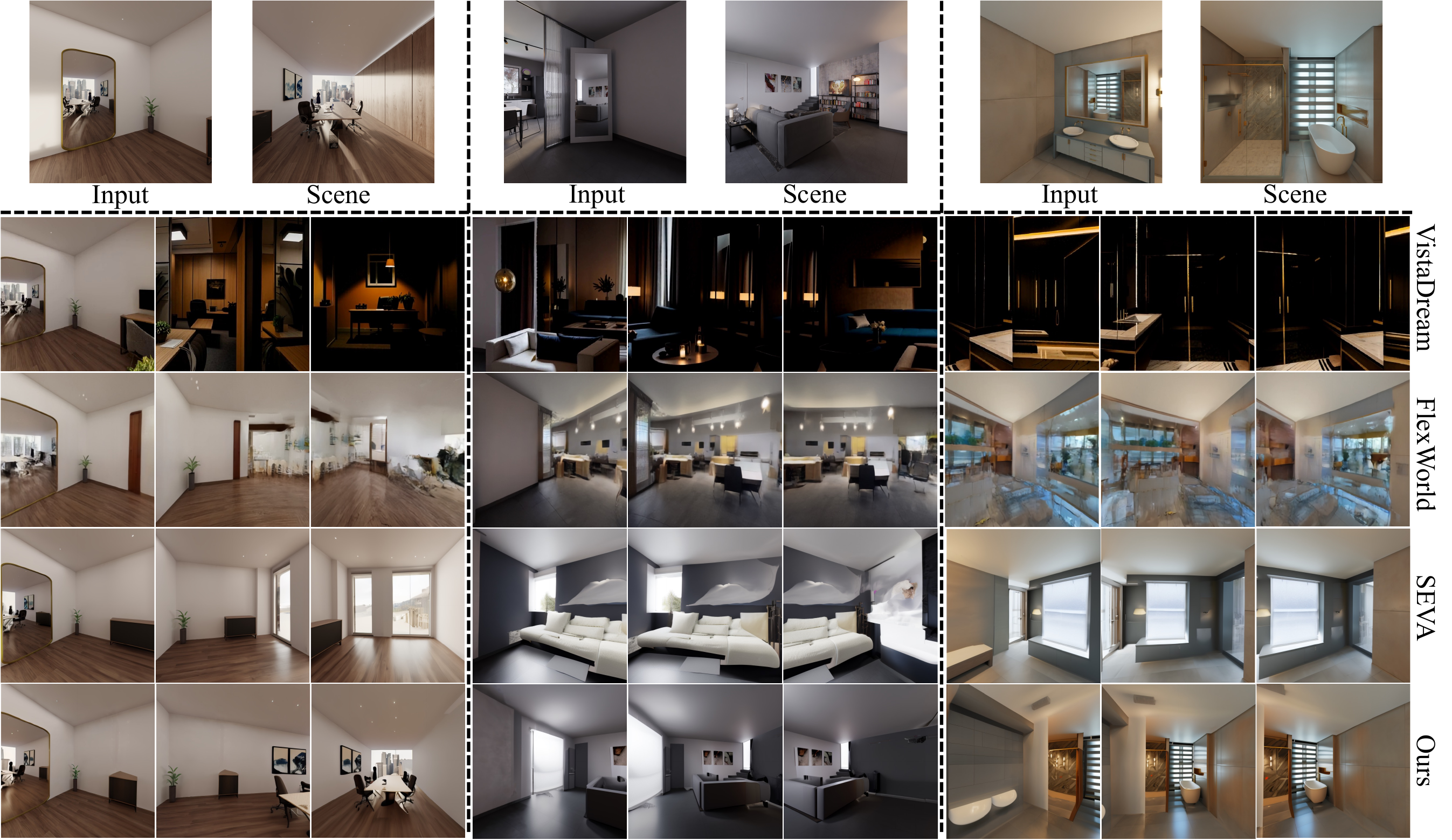}
    \caption{Qualitative comparison in the single image NVS setting. Despite observing only single input image, \nickname utilizes mirror cues to generate coherent novel views, outperforming prior methods~\cite{wang2025vistadream, chen2025flexworld, zhou2025stable} on reflected and hidden regions.}
    \label{fig:single_view_qual}
\end{figure}

\paragraph{Evaluation protocol in single image input setting}
The single image NVS evaluation is affected by scale and pose ambiguity, making per-view pixel metrics not directly comparable to ground truth. 
We therefore report DreamSim and CLIP similarity to assess perceptual alignment between ground truth and generated images, which can be seen as approximately scale-invariant. 
For each predicted target, we compute two perceptual scores for each metric: (i) relative to the input image and (ii) relative to a ground-truth image showing the reflected region along the test trajectory (as shown in the first row of Fig.~\ref{fig:single_view_qual}). 
We average these two scores to capture both global coherence with the input and semantic alignment with the reflected content. 

\setlength{\intextsep}{0pt}        
\setlength{\columnsep}{6pt}        

\begin{wraptable}{r}{0.38\linewidth}
\vspace{-14pt}
\centering
\footnotesize
\caption{Single image result.}
\begin{tabular}{ccc}
\hline
Method & DS$\downarrow$ & CLIP$\uparrow$ \\ \hline
VistaDream~\cite{wang2025vistadream} & 0.569 & 0.866 \\
FlexWorld~\cite{chen2025flexworld}   & 0.369 & 0.879 \\
SEVA~\cite{zhou2025stable}           & 0.377 & 0.905 \\ 
\hline
Ours                                 &\cellcolor{pastelblue}0.341 &\cellcolor{pastelblue}0.911 \\ \hline
\end{tabular}
\label{tab:single-quant-table}
\end{wraptable}
\paragraph{Results}
Table~\ref{tab:single-quant-table} shows that \nickname performs better than VistaDream~\cite{wang2025vistadream}, FlexWorld~\cite{chen2025flexworld}, and SEVA in terms of DreamSim (DS) and CLIP similarity with the input view and the ground truth reflected view, suggesting that incorporating explicit reflection reasoning is beneficial even when only a single image is available.
Figure~\ref{fig:single_view_qual} shows that prior methods often struggle to interpret mirror-reflected regions from limited evidence, leading to incomplete or less coherent predictions when synthesizing the target view. 
\nickname, by contrast, leveraging the reflected information in the input image and produces novel views with more aligning the scene context and stable structures in the mirror-revealed areas.


\begin{table}[t]
\centering
\caption{Ablation study on injection of reflected virtual view, Mirror-gated attention (Mirror-gated attn.), and two-step generation process (Two-step gen.).
We compare the performance by adding each component on the baseline model, SEVA~\cite{zhou2025stable}.
}

\begin{tabular}{ccc|ccccc}
\hline
Virtual view  & Mirror-gated attn. & Two-step gen. & DS $\downarrow$ & CLIP$\uparrow$ &PSNR$\uparrow$ & SSIM$\uparrow$ & LPIPS$\downarrow$  \\
\hline

X & X & X & 0.270 & 0.924 & 13.62 & 0.612 & 0.492  \\
O & X & X & 0.229 & 0.923 &  13.59  & 0.643 & 0.536  \\
O & O & X & \cellcolor{lightblue}0.160 & \cellcolor{pastelblue}0.948 &  \cellcolor{lightblue}15.01  & \cellcolor{lightblue}0.671 & \cellcolor{lightblue}0.416  \\
O & O & O & \cellcolor{pastelblue}0.156 & \cellcolor{pastelblue}0.948 & \cellcolor{pastelblue}15.27 & \cellcolor{pastelblue}0.676 &\cellcolor{pastelblue}0.403  \\

\hline
\end{tabular}

\label{tab:ablationstudy}
\end{table}
\begin{figure*}[!t]
    \centering
        \includegraphics[width=1\linewidth]{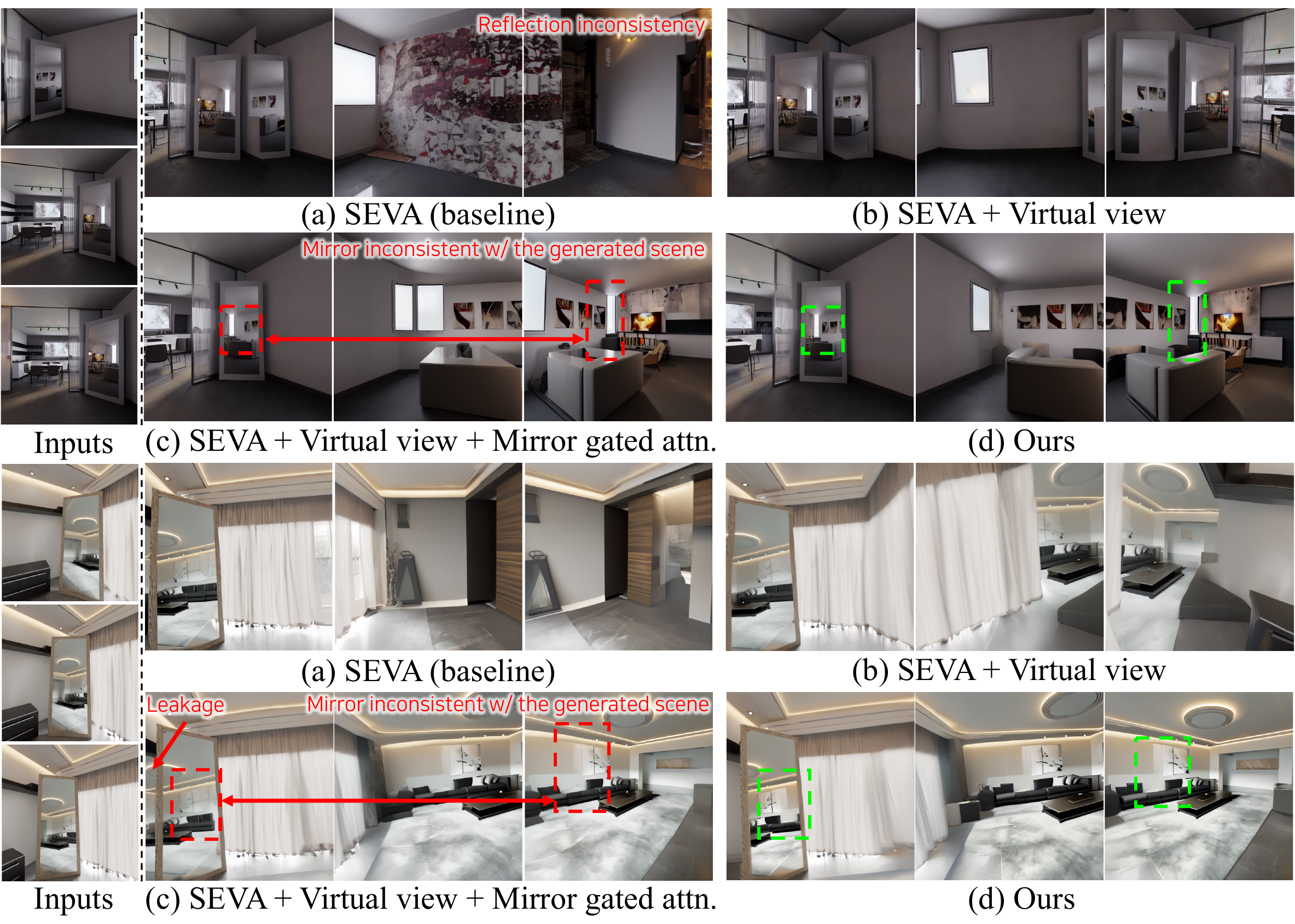}
    \caption{
Ablation on synthetic scenes by progressively adding our components: (a) SEVA baseline, (b) + virtual view, (c) + Mirror-gated attention, and (d) + two-stage generation (full method). Our full model produces reflections consistent with the generated scene.
    }
    \label{fig:ablation_qualitative_supp}
\end{figure*}
\subsection{Ablation Study}
\label{sec:ablation_study}

We conduct ablation studies on the synthetic dataset under the sparse-input setting. 
As shown in Fig.~\ref{fig:ablation_qualitative_supp}, we progressively add each component to the baseline SEVA~\cite{zhou2025stable}: 
(a) the baseline SEVA, (b) + reflected virtual view, (c) + Mirror-gated attention, and (d) + two-step generation (full method).
The baseline often produces scenes inconsistent with the reflection observed in the mirror. 
Introducing the virtual view improves geometric reasoning for reflections but still allows non-reflected regions in the inputs to influence the mirror region. 
Adding Mirror-gated attention restricts the attention to reflection-relevant regions, preventing leakage from non-mirror areas and reducing duplicated structures or artifacts.
Finally, the two-step generation further improves the alignment between the generated scene and the mirror surface. 
Although this improvement is not strongly reflected in the quantitative metrics, it can be clearly observed in the results, where the reflected scene becomes better aligned with the generated scene.
Table~\ref{tab:ablationstudy} also shows that adding components leads to performance improvements.



\subsection{Applications}
\begin{figure}[!t]
    \centering
        \includegraphics[width=1.\linewidth]{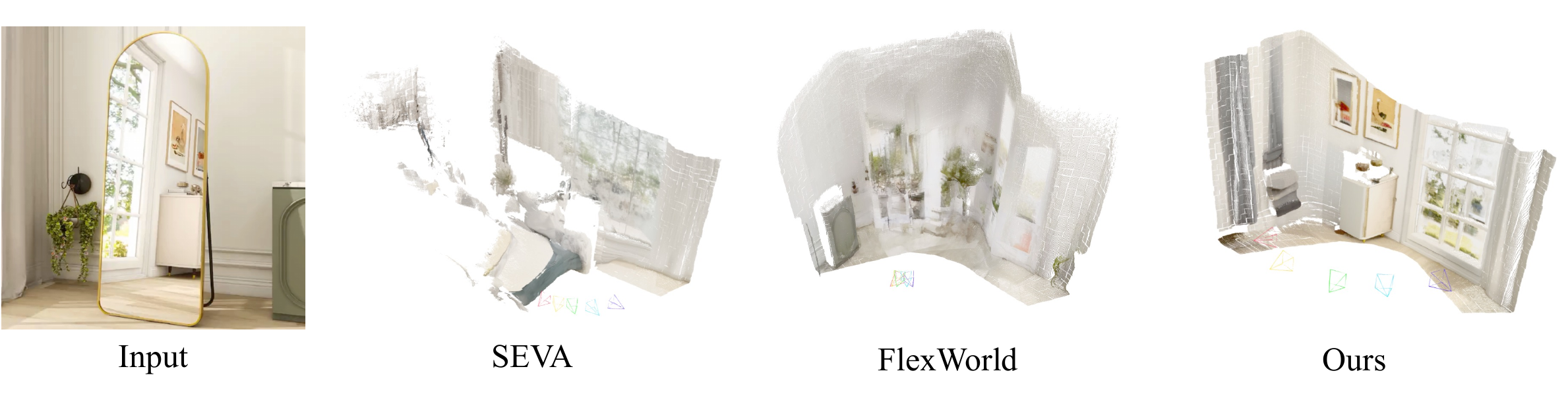}
    \caption{Qualitative comparison of 3D reconstruction with generated images using $\text{Pi}^3$~\cite{wang2025pi}. \nickname produces more complete and stable geometry in mirror-containing scenes compared to FlexWorld~\cite{chen2025flexworld} and SEVA~\cite{zhou2025stable}, with a single input image.}
    \label{fig:3d_recon_qual}
\end{figure}
\begin{figure}[!t]
    \centering
    \includegraphics[width=1.0\linewidth]{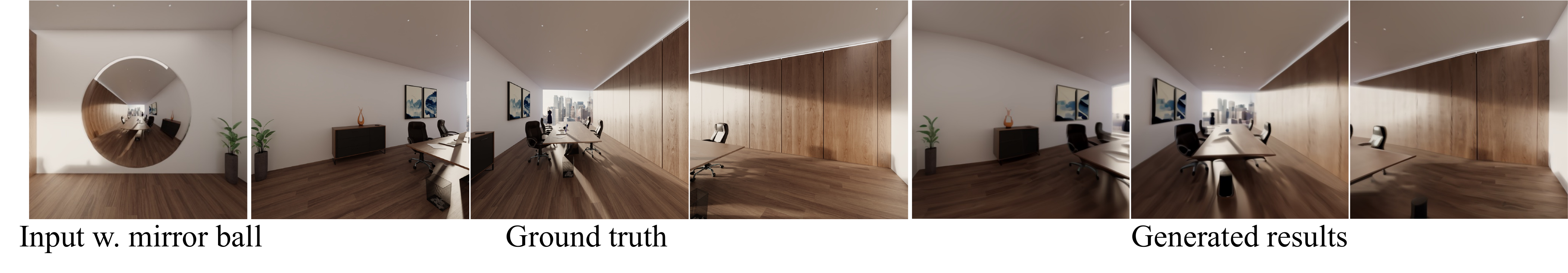}
    \vspace{-12pt}
    \caption{Result using an input with a distorted mirror. We warp the distorted reflection into the planar-mirror representation and apply
Ref-GeNVS without modification.
}
\label{fig:rebut_distortion}
\end{figure}
\paragraph{3D reconstruction} We compare the 3D reconstructions obtained by feeding the generated novel views from each model into Pi³~\cite{wang2025pi}. Since Pi³ relies heavily on the 3D geometry of input views, errors in mirror-revealed regions directly show up as broken geometry. As shown in \Fref{fig:3d_recon_qual}, our results show better alignment between reflected content, leading to reflection-consistent reconstructions.

\paragraph{Generalization to distorted reflections}
To explore an extension to distorted reflections, we conduct an experiment with a distorted mirror by constructing a scene containing a mirror ball. Using the given mirror ball geometry, we warp the distorted reflection into the planar-mirror representation and apply Ref-GeNVS until Stage 1 without changing the pipeline.
Figure~\ref{fig:rebut_distortion} shows that our method can use wide-field reflections from distorted mirrors, such as convex mirrors, demonstrating broader practical utility.
\section{Conclusion}
\label{sec:conclusion}
We propose \nickname, a reflection-aware, training-free novel view synthesis method that augments a multi-view diffusion backbone with reflected views. 
By estimating the mirror plane and reflecting camera poses, our method converts mirror-containing inputs into complementary viewpoints that guide generation toward reflection-consistent results in a training-free manner. Empirically, \nickname synthesizes novel views that better align with mirror-reflected content and expose global scene structure compared to prior approaches.
Our work has the potential to benefit a wide range of applications, including 3D scene generation and embodied agents. For example, by leveraging partial mirror observations, our method can enable embodied agents to infer and predict regions beyond their direct field of view, facilitating more informed navigation.


\paragraph{Future work}
As noted in Sec.~\ref{sec:intro}, reflection-aware generative NVS is a physically constrained generation problem, and na\"ively training a model without explicit physical priors is not guaranteed to learn reflection constraints implicitly. Given this, extending our approach to a trainable framework by incorporating mirror detection and reflected-view conditioning during training is a promising direction for future work, as it could provide explicit physical priors for mirror scenes.

\vfill
{\footnotesize\paragraph{Acknowledgements}
This work was supported by Institute of Information \& Communications Technology Planning \& Evaluation (IITP) grant (No.RS-2025-25443318, Physically-grounded Intelligence: A Dual Competency Approach to Embodied AGI through Constructing and Reasoning in the Real World; No.RS-2025-02263036, Generative AI-based Pre-visualization Technology for Media Production Coordination), the National Research Foundation of Korea (NRF) grant (No. RS-2024-00358135, Corner Vision: Learning to Look Around the Corner through Multi-modal Signals), and the InnoCORE program (26-InnoCORE-01) funded by the Korea government (MSIT).
\par}


%
%
\bibliographystyle{splncs04}
\bibliography{references}
\end{document}


\title{Reflection-aware Generative Novel View Synthesis \\
\vspace{10pt}
\large{Supplementary Material}}


\titlerunning{Ref-GeNVS}

\author{GeonU Kim\orcidlink{0009-0009-0224-0060} \and
Shin Dong-Yeon\orcidlink{0009-0000-1764-7829} \and
Tae-Hyun Oh\orcidlink{0000-0003-0468-1571}
}

\authorrunning{Kim et al.}

\institute{KAIST\\
\email{\{geonukim,shindy,taehyun.oh\}@kaist.ac.kr}}

\maketitle

\section*{Contents}
 \begin{itemize}
    \item \Sref{sec:supp_implementation_details}. Implementation Details.
    \item \Sref{sec:robustness}. Robustness on inaccurate mirror detection.
    \item \Sref{sec:supp_limitations}. Limitations.
    \item \Sref{sec:supp_add_exp.}. Additional experimental results.
\end{itemize}

{
\hypersetup{linkcolor=black}
\tablesupcontents
}

\setcounter{section}{0}
\renewcommand{\thesection}{\Alph{section}}
\renewcommand{\thesubsection}{\thesection.\arabic{subsection}}
\renewcommand{\thefigure}{S\arabic{figure}}
\renewcommand{\thetable}{S\arabic{table}}

\section{Implementation details}
\label{sec:supp_implementation_details}
This section provides additional implementation details for Stage~1 (Sec.~\ref{sec:supp_stage1}), Stage~2 (Sec.~\ref{sec:supp_stage2}), automation pipeline for preprocess (Sec.~\ref{sec:supp_preprocess}), and experimental details (Sec.~\ref{sec:supp_experment}).

\subsection{Stage 1: Mirror-gated attention}
\label{sec:supp_stage1}
Stage~1 generates masked target views $\mathbf{I}^{\mathrm{mask}}_o$ using mirror-masked original inputs $(\mathbf{I}^{\mathrm{mask}}_s,\mathbf{P}_s)$ and reflected inputs $(\mathbf{I}'^m_s,\mathbf{P}'^m_s,\mathbf{M}'^m_s)$.
The reflected mirror mask $\mathbf{M}'^m_s$ is defined in the image space and must be aligned with the token layout of the attention module. We downsample $\mathbf{M}'^m_s$ to the spatial resolution of the attention feature map and reshape it into a token-level binary mask $m \in \{0,1\}^{B \times N}$, where $B$ is the batch size and $N$ is the number of tokens per view. The mask is then broadcast to match the attention tensor layout used in the transformer blocks.

\paragraph{Padding-based masked attention}
Mirror-gated attention is implemented using a padding trick compatible with the fused scaled dot-product attention operator in PyTorch~\cite{paszke2019pytorch} without custom kernels. Let $Q,K,V \in \mathbb{R}^{B \times H \times N \times D}$ denote the query, key, and value tensors, where $H$ is the number of attention heads and $D$ is the head dimension. Instead of explicitly applying an attention mask, we append an additional padding channel to the attention inputs.

Specifically, we construct padded tensors $Q'=[Q,\mathbf{1}]$, $K'=[K,b]$, and $V'=[V,\mathbf{1}]$, resulting in tensors of shape $\mathbb{R}^{B \times H \times N \times (D+1)}$. Here $\mathbf{1}$ denotes a padding channel filled with ones and $b \in \mathbb{R}^{B \times H \times N \times 1}$ encodes the mirror mask such that $b=0$ for tokens inside the mirror region and $b=-C$ for tokens outside the mirror region, where $C$ is a large constant (we use $C=10^4$ in our implementation). This effectively introduces a large negative bias to the attention logits corresponding to non-mirror tokens.

The attention output is then computed using the fused scaled dot-product attention operator with the FlashAttention~\cite{dao2022flashattention} backend as: 

\[O'=\mathrm{FlashAttention}(Q',K',V')\]. 

Finally, the appended padding channel is removed by slicing the last feature dimension, producing the output $O \in \mathbb{R}^{B \times H \times N \times D}$.
Note that this formulation allows Mirror-gated attention to be incorporated directly into the attention computation while remaining fully compatible with the optimized FlashAttention implementation, \emph{without custom kernels}.

Mirror-gated attention is applied to all cross-view attention layers in the multi-view diffusion backbone, including both the 3D attention and the 1D attention modules~\cite{zhou2025stable}.
Mirror-gated attention is also applied to 2D self-attention on the reflected virtual views to restrict receptive fields on the reflected region, ensuring that only reflected regions contribute to the target-view generation. 
To further mitigate artifacts from non-mirror region, we inpaint pixels around the mirror boundary in the reflected views using an off-the-shelf inpainting model~\cite{suvorov2021resolution} before being used as inputs to the diffusion backbone. We also use self-refining sampling~\cite{jang2026self} for more stable multi-view generation.

\subsection{Stage 2: Reflection injection}
\label{sec:supp_stage2}


\paragraph{Input condition and Mirror-gated attention in Stage~2}
Stage~2 uses the same multi-view diffusion backbone including Mirror-gated attention as Stage~1 and conditions on the original views $(\mathbf{I}_s,\mathbf{P}_s)$ and the reflected inputs $(\mathbf{I}'^{\,m}_s,\mathbf{P}'^{\,m}_s,\mathbf{M}'^{\,m}_s)$. 
In Stage 2, since our goal is to synthesize the mirror pixels in $\mathbf{I}_o^{\text{mask}\_m}$, the mirror region $\mathbf{M}_o^m \mathbf{I}_o^{\text{mask}\_m}$ should not influence the generation process. Therefore, 
Stage~1 output $\mathbf{I}_o^{\text{mask}\_m}$ is not used as a source view for conditioning, instead, only the non-mirror region of $\mathbf{I}_o^{\text{mask}\_m}$ is used to guide the diffusion process through the SDEdit-style~\cite{meng2022sdedit} guidance described below.

\paragraph{SDEdit-style guidance parameters}
For the SDEdit-style guidance, we initialize the diffusion process from an intermediate noisy latent obtained by forward diffusing the encoded Stage~1 output $E(\mathbf{I}_o^{\text{mask}\_m})$. 
Since most regions of the image are already determined by conditioning inputs and only a small boundary around the mirror needs to be synthesized, we give guidance until $t_0=42$ out of the total diffusion steps $T=50$. 
This reduces unnecessary resampling of the surrounding context while allowing sufficient stochasticity to generate natural mirror boundaries.

To avoid abrupt transitions between the injected mirror latent and the surrounding region, we use a soft mirror mask $\tilde{\mathbf{M}}_o^{m}$ obtained by applying Gaussian blur to the binary mask before resizing it to the latent resolution. 
This soft mask enables smooth blending between the generated mirror region and the surrounding scene during the early denoising steps.
After timestep $t_0$, the guidance is removed and the diffusion process proceeds normally so that the diffusion prior can naturally harmonize the mirror boundary.

\begin{figure*}[!t]
    \centering
    \includegraphics[width=1.0\linewidth]{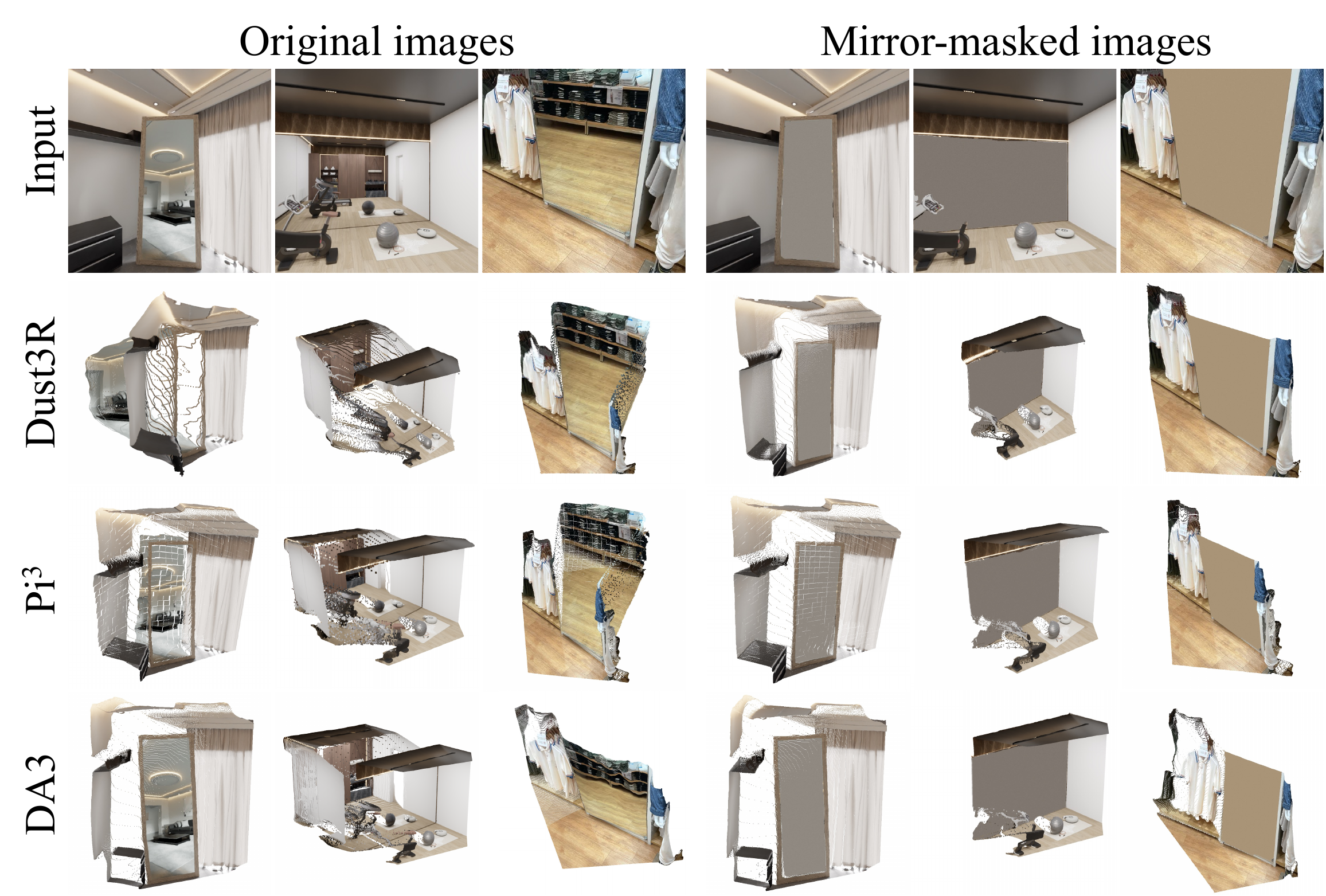}
    \caption{
    3D point prediction with Dust3R~\cite{wang2024dust3r}, $\text{Pi}^3$~\cite{wang2025pi}, and Depth Anything 3 (DA3)~\cite{lin2025depthanything3} on original images (left) and mirror-masked images (right). $\text{Pi}^3$ and DA3 randomly predict 3D points of mirror surface according to the input image: as flat plane or as a hole in the scene geometry. After masking the mirror, they consistently predict the mirror as a flat plane.
    }
    \label{fig:mirror_masked_depth}
\end{figure*}
\subsection{Automation pipeline for preprocess}
\label{sec:supp_preprocess}
Estimating mirror geometry is challenging since reflections violate multi-view photometric consistency on the mirror surface, which hinders 3D reconstruction methods that rely on multi-view consistency \cite{wang2025vggt, wang2024dust3r, wang2025pi, schoenberger2016sfm, lin2025depthanything3}. 
MirrorGaussian~\cite{liu2024mirrorgaussian}, a prior work, estimates the mirror plane using SfM~\cite{schoenberger2016sfm} points located on the mirror edge with dense input setting (more than 30 input images). However, this approach becomes unstable in our setting with extremely sparse views (from 1 to 3) where reliable edge correspondences are difficult to obtain using traditional 3D reconstruction method, such as SFM~\cite{schoenberger2016sfm}. Edge-based mirror plane estimation a often fails when the mirror is occluded or the segmentation mask is inaccurate (see Sec.~\ref{sec:robustness}).

To estimate mirror geometry under sparse-view inputs, we adopt recent feed-forward 3D regression methods~\cite{wang2024dust3r,wang2025pi, lin2025depthanything3} that predict per-pixel 3D points, which are more applicable in sparse-view settings than SFM. 
However, these methods exhibit unstable behavior on mirror regions due to the inherent ambiguity of reflections. 
Depending on the scene content, the mirror area may be predicted either as a flat plane or as geometry behind the mirror (\textit{i.e.}, perceived as a hole in the scene). 
For example, as shown in Figure~\ref{fig:mirror_masked_depth}, Dust3R tends to interpret the mirror as a hole, while Pi3 and Depth Anything 3 (DA3) produce inconsistent predictions, alternating between planar surfaces and holes. This ambiguity hinders robust mirror plane estimation, especially when the mirror mask is imperfect and cannot reliably extract accurate 3D edge points of mirror from the obtained 3D points.   

To mitigate this ambiguity, we first mask the mirror regions in the input images using a single color (the average value of $\textbf{I}_s^m$ in our experiments) before performing 3D prediction. 
Specifically, we estimate binary mirror masks $\mathbf{M}_s^m$ from the input images with mirrors $\mathbf{I}_s^m$ using DAM~\cite{xing2023DAM}. 
The masked images are then fed into a per-pixel 3D point regression model to predict 3D point maps $\mathbf{X}_s$. 
Figure~\ref{fig:mirror_masked_depth} shows that mirror masking enables consistent depth estimation on the mirror surface across all methods, allowing the mirror to be reconstructed as a planar surface. 
We use Depth Anything 3~\cite{lin2025depthanything3} in our experiments.

\paragraph{Pose-free case}
When camera poses are not provided, we first predict per-pixel 3D points $\textbf{X}_s$ and camera poses $\textbf{P}_s$ from $\textbf{I}_s$ using a 3D point regression method Depth~\cite{wang2025pi}. 
To prevent reflection-induced geometry errors inside the mirror, we mask the mirror regions in $\textbf{I}_s$ with a single color before performing the 3D prediction. 
From the predicted 3D point maps $\textbf{X}_s$, we extract the points corresponding to mirror pixels to obtain surface points $\textbf{X}_s^{\text{mirror}}$. 
The mirror plane equation $\mathbf{e}=[\mathbf{n},d]$ is then estimated by fitting a plane to these surface points using the RANSAC algorithm~\cite{fischler1981random}.

\paragraph{Given-pose case}
When camera poses $\textbf{P}_s$ are provided, the poses predicted by DA3 may lie in a different coordinate system from the given poses. 
To resolve this inconsistency, we align the predicted camera poses to the given poses using Umeyama alignment. 
Specifically, we extract camera centers from both pose sets and estimate a similarity transform consisting of rotation, translation, and global scale. 
The recovered scale is used to rescale the predicted depth, after which the mirror surface points $\textbf{X}_s^{\text{mirror}}$ are used to estimate the mirror plane $\mathbf{e}=[\mathbf{n},d]$ via RANSAC.

\paragraph{Single-image case}
The single-image case also introduce a scale mismatch since the 3D regression model predict geometry in its own scale, while the generative backbone, \eg, SEVA~\cite{zhou2025stable}, operates in the scale induced by its multi-view diffusion prior. 
To calibrate the two different scales, we begin by masking the mirror in the input, which avoids artifacts (see Fig.~1) during generation, then use generative backbone to generate a stereo image with a small parallel camera translation.
We feed this image pair into the 3D regression model to estimate 3D points and the relative camera poses. The resulting two images and relative poses are then used in \nickname as if they were sparse multi-view inputs, which calibrates the mirror-plane estimation to the backbone’s scale without external calibration.

\subsection{Experimental details}
\label{sec:supp_experment}
All experiments were conducted on a single NVIDIA RTX 6000 Ada Generation GPU. 
The image resolution was set according to the recommended input resolution of each competing method. 
For instance, MVGenMaster~\cite{cao2025mvgenmaster} and our method based on MVGenMaster operate at $512\times512$ resolution, while SEVA~\cite{zhou2025stable} and our method based on SEVA use $576\times576$ resolution.
We use a real-world Mirror-NeRF dataset~\cite{zeng2023mirror-nerf} and our synthetic dataset for our evaluation. The Mirror-NeRF dataset consists of three scenes, and our synthetic dataset consists of eight scenes.
For sparse-view novel view synthesis, we use three input images and evaluate the generation on nine target views per scene. 
For single-image novel view synthesis, we use one input image and evaluate the generation on fourteen target views per scene.
We use our automated preprocess on the synthetic dataset, while use ground truth mirror mask and mirror plane on the Mirror-NeRF dataset.

\begin{figure*}[!t]
    \centering
    \includegraphics[width=1.0\linewidth]{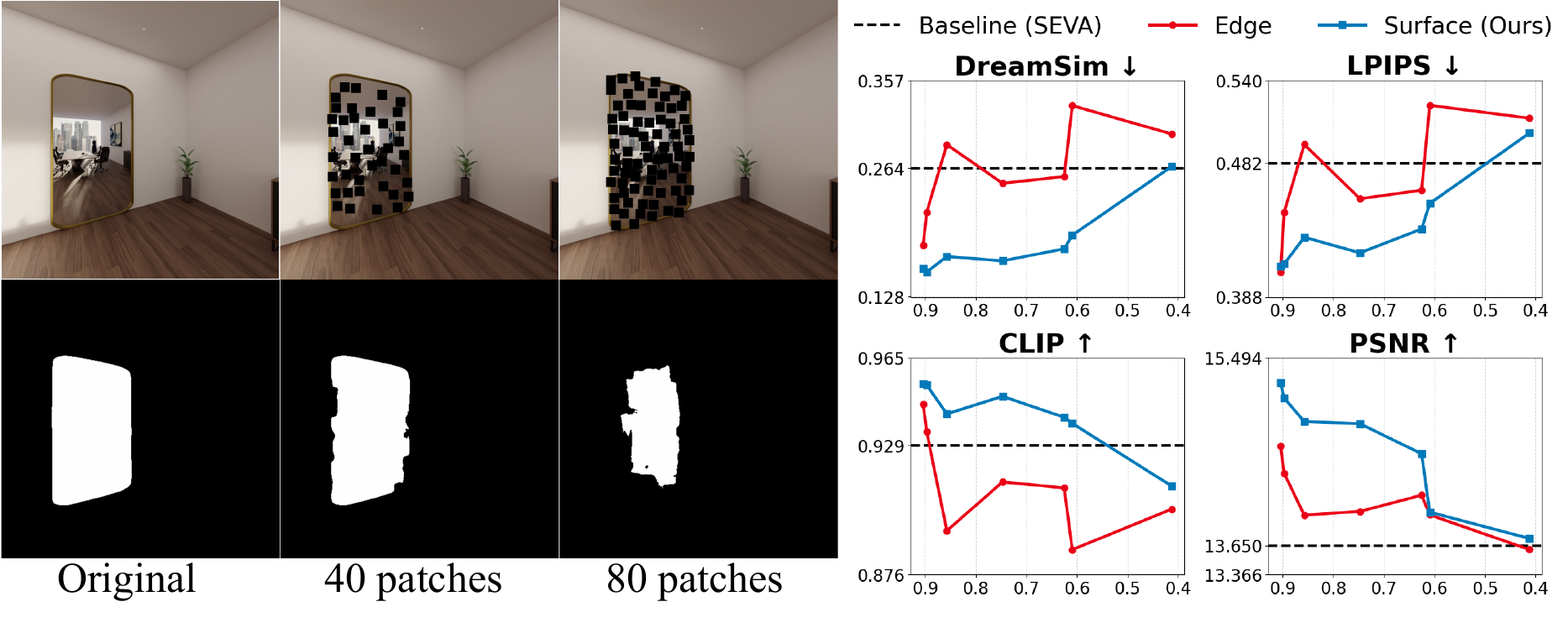}
    \caption{
    Robustness analysis under imperfect mirror masks.
    We corrupt the mirror region by pasting random patches before DAM-based mirror segmentation and measure the IoU between the predicted mirror mask and the ground-truth mirror mask.
    The x-axis shows the mask IoU and the y-axis shows reconstruction performance measured by DreamSim, LPIPS, CLIP similarity, and PSNR.
    We compare two mirror plane estimation strategies: an edge-based method (Edge) that fits the mirror plane from edge points without mirror masking, and our surface-based method (Surface) that applies mirror masking and estimates the plane from surface points predicted by per-pixel 3D regression.
    As mask IoU decreases, the performance of both methods degrades, but the proposed surface-based estimation remains consistently more robust than the edge-based approach.
    The dashed line indicates the baseline NVS model (SEVA).
    }
    \label{fig:robustness}
\end{figure*}
\begin{figure*}[!t]
    \centering
    \includegraphics[width=1.0\linewidth]{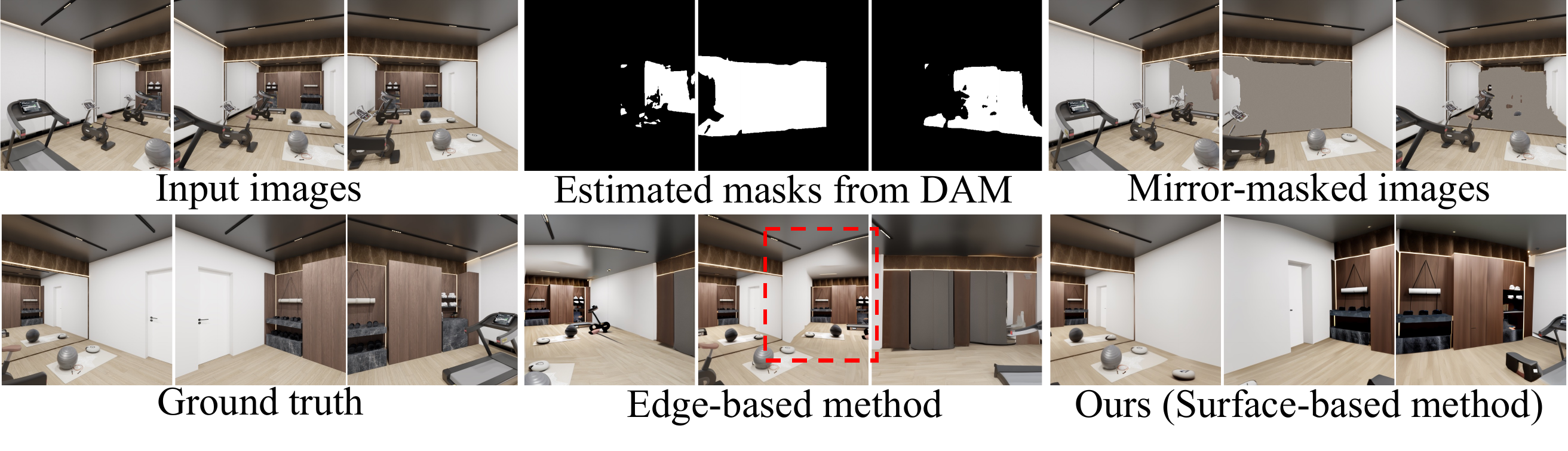}
    \caption{
        Failure case of edge-based mirror plane estimation under imperfect mirror masks.
        Given input images, mirror masks estimated by DAM contain inaccuracies.
        When mirror plane estimation relies on edge points, the inaccurate mask leads to incorrect plane estimation and misaligned reflected content.
        In contrast, our surface-based mirror plane estimation remains robust by fitting the plane using surface points predicted from per-pixel 3D regression, producing reflection-consistent novel views.
    }
    \label{fig:robust_case}
\end{figure*}
\section{Robustness on inaccurate mirror detection}
\label{sec:robustness}
We assess the robustness of the mirror plane estimation under imperfect mirror masks by artificially corrupting the mirror region before mirror detection, on our synthetic dataset. 
Specifically, instead of directly corrupting the mirror masks, we paste random patches on the mirror area before running DAM-based~\cite{xing2023DAM} segmentation to simulate realistic mirror detection errors. 
By progressively increasing the number of pasted patches, we gradually degrade the quality of the detected mirror masks and measure the resulting mask IoU.
We then proceed to the remaining preprocess, Stage 1 and Stage 2 with the degraded mirror masks and evaluate the performance, DreamSim~\cite{fu2023dreamsim}, CLIP similarity~\cite{hessel2021clipscore}, PSNR and LPIPS~\cite{zhang2018lpips}.
For each patch corruption level, we measure the IoU between the predicted mirror mask and the ground-truth mirror mask and analyze the correlation between mask IoU and reconstruction performance.

We compare two mirror plane estimation strategies: 
(i) an edge-based approach that estimates the mirror plane from edge points without mirror masking, and 
(ii) our surface-based approach that first applies mirror masking and then fits the mirror plane using surface points predicted by per-pixel 3D regression. 
As the mask IoU decreases, the performance of both methods degrades; however, the surface-based estimation remains significantly more robust than the edge-based method, as shown in Fig.~\ref{fig:robustness}. Figure~\ref{fig:robust_case} also shows a failure case of edge-based method with imperfect mirror mask and shows robustness of our method.

Notably, even when the mask IoU drops to around 0.6, our surface-based method still outperforms the baseline, SEVA~\cite{zhou2025stable}. 
This demonstrates that the proposed mirror-masking and surface-point-based estimation provides a more stable mirror plane estimation under imperfect mirror detection.
Note that prior mirror-related NVS methods (\eg, Mirror-NeRF~\cite{zeng2023mirror-nerf}, MirrorGaussian~\cite{liu2024mirrorgaussian}) assume given mirror masks, however, we aim to relax this strong assumption by introducing a mask-tolerant mirror plane estimation pipeline.

\begin{figure}[!t]
    \centering
        \includegraphics[width=0.95\linewidth]{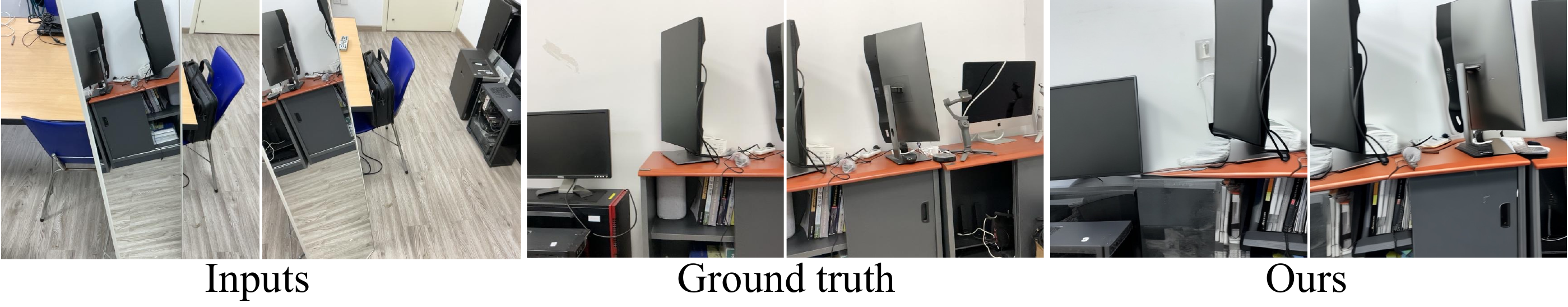}
    \caption{
    Failure case. In extreme cases where point cloud regression is unreliable (\eg the mirror is only partially visible), inaccurate mirror-plane estimates can produce results that are misaligned with the ground truth at the same camera poses.
    }
    \label{fig:limitation}
\end{figure}
\section{Limitations}
\label{sec:supp_limitations}

Mirror plane estimation becomes difficult in extreme cases where boundary correspondences are scarce or unstable, and the 3D point regression method~\cite{lin2025depthanything3} may fail to recover an accurate mirror plane.
This occurs when the mirror occupies a very small image region or appears only partially. 
Since our automated preprocess depends on Depth Anything 3, failures or instability in these cues can lead to misalignment of the reflected content in the final generation, as shown in Fig.~\ref{fig:limitation}. 
Nonetheless, since our model can utilize any 3D point regression method, the performance of our method stands to benefit as advanced models emerge.
This limitation is consistent with the robustness analysis presented earlier. 
As shown in Fig.~\ref{fig:robustness}, our mask-tolerant mirror plane estimation pipeline remains relatively stable under imperfect mirror masks, even when the mask IoU significantly decreases. 
However, in extreme cases where reliable geometric cues are insufficient, such as very small or heavily occluded mirrors, accurate mirror plane estimation may still fail when the 3D regression method fails. 
These results suggest that while our pipeline improves robustness to realistic mask errors, extremely challenging geometric conditions remain a limitation of the current preprocessing pipeline and influence final generation results.

\begin{figure*}[!t]
    \centering
        \includegraphics[width=1.0\linewidth]{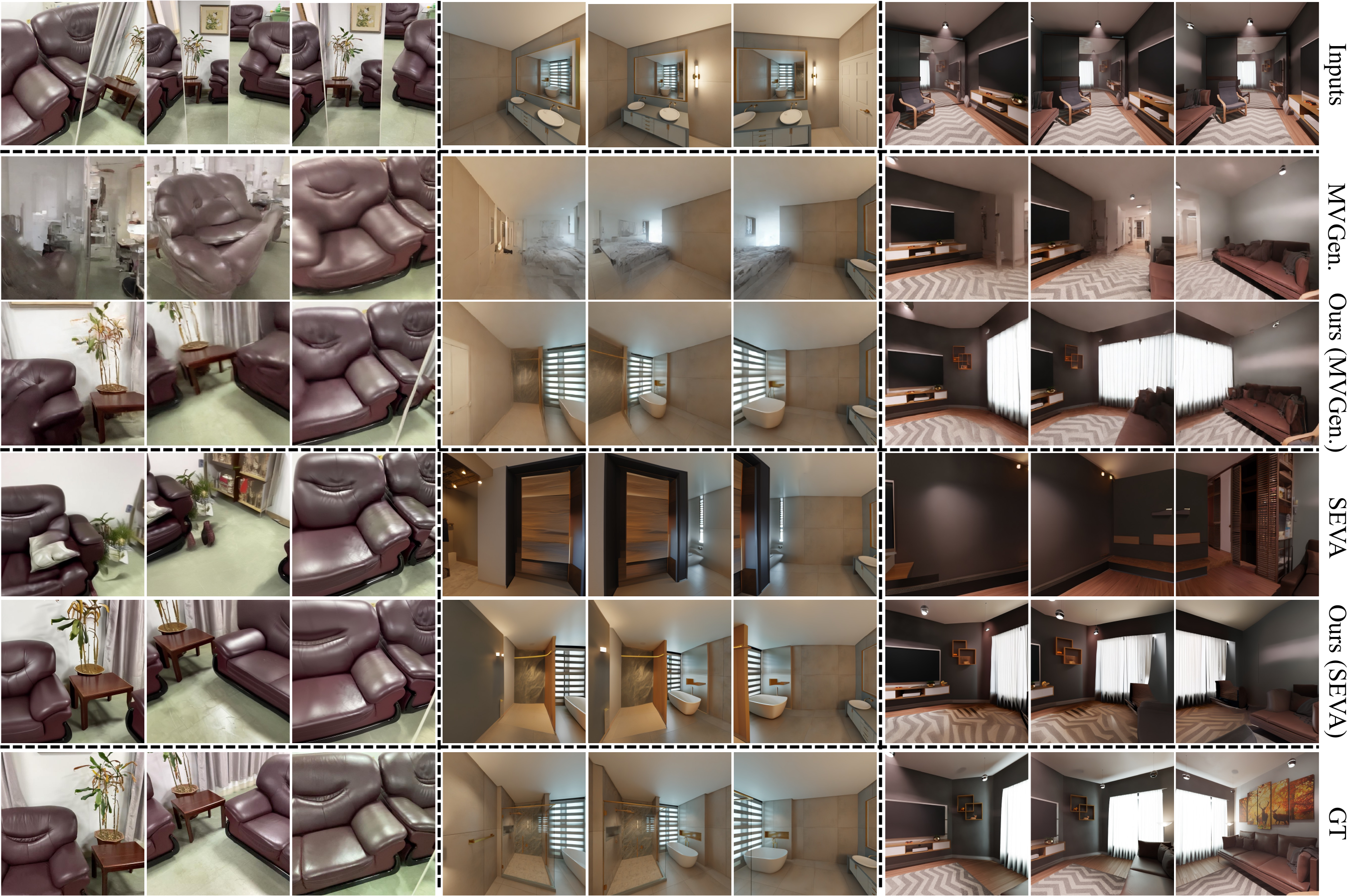}
    \caption{
    Additional qualitative comparison of sparse images novel view synthesis with the prior methods~\cite{zhou2025stable, cao2025mvgenmaster}. The left-most scene is from the real dataset~\cite{zeng2023mirror-nerf} and the middle and right scenes are from our synthetic dataset.
    }
    \label{fig:supp_multiview_qual}
\end{figure*}
\begin{figure*}[!t]
    \centering
        \includegraphics[width=1.0\linewidth]{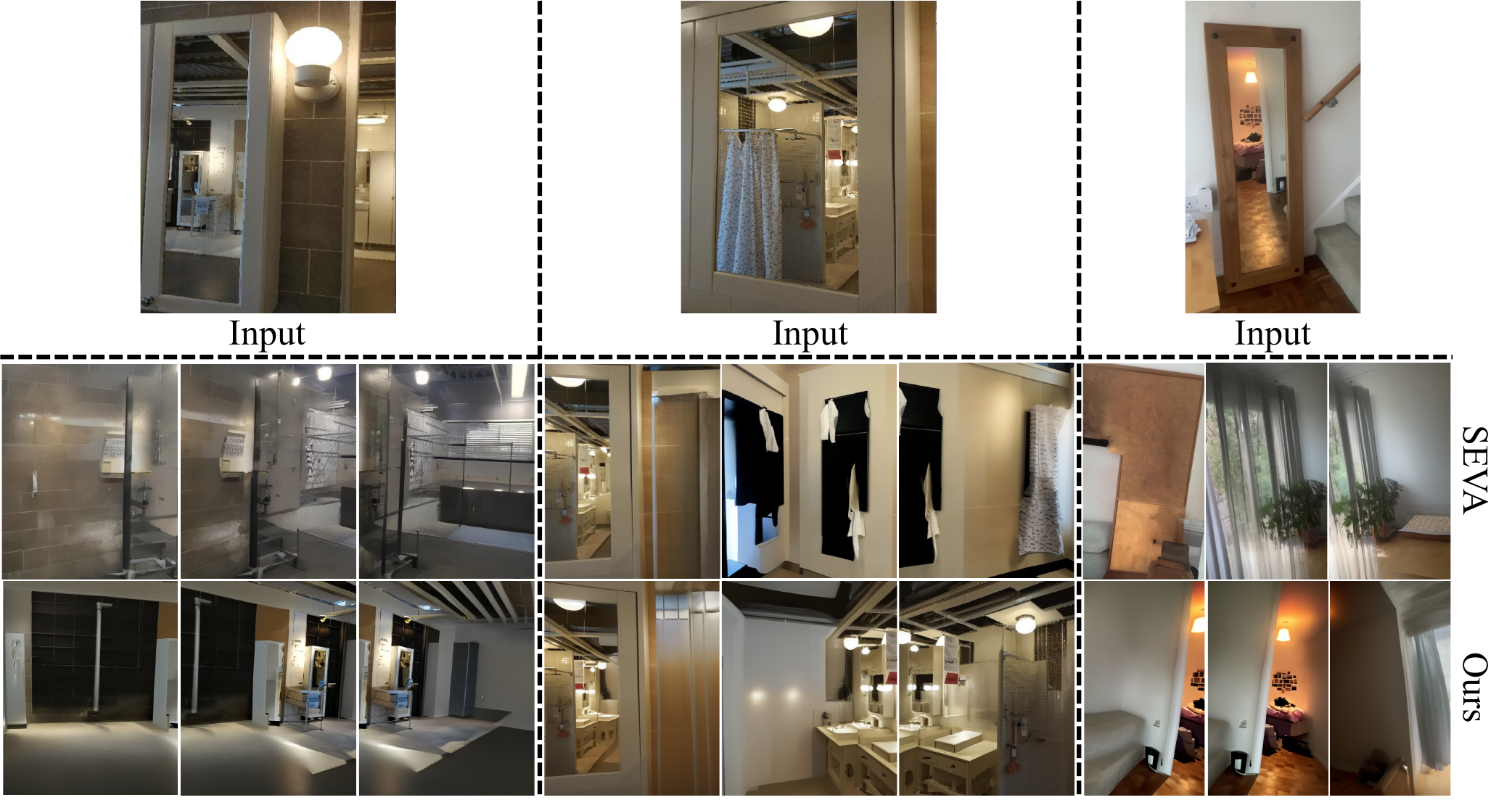}
    \caption{
    Additional qualitative results of single image novel view synthesis on the real-world dataset, the mirror segmentation dataset (MSD)~\cite{yang2019my} (left and middle) and the MMD dataset~\cite{warren2024effective} (right). There is no ground truth scene images since the datasets are originally designed for mirror segmentation evaluation.
    }
    \label{fig:supp_singleview_qual}
\end{figure*}

\section{Additional experimental results}
\label{sec:supp_add_exp.}
In this section, we provide additional experimental results including qualitative results and infernece time.

\paragraph{Sparse images novel view synthesis}
We provide additional qualitative comparisons of sparse images novel view synthesis
on both the real dataset~\cite{zeng2023mirror-nerf} and our synthetic dataset. 
As shown in Fig.~\ref{fig:supp_multiview_qual}, prior methods~\cite{cao2025mvgenmaster,zhou2025stable} fail to generate reflection-consistent scene,
 in contrast, \nickname generates reflected content consistent with the scene structure seen on the mirror. 
 
\paragraph{Single image novel view synthesis}
We provide additional single-image novel view synthesis results on real-world datasets, including the MMD dataset~\cite{warren2024effective} and the mirror segmentation dataset (MSD)~\cite{yang2019my}. 
Since these datasets are originally designed for mirror segmentation evaluation, they do not provide ground-truth images of the scene behind the mirror for novel view synthesis. 
We use 360-degree orbit target camera poses around the scene as conditions to generate a trajectory of novel views from a single input. 
As shown in Fig.~\ref{fig:supp_singleview_qual}, SEVA struggles to exploit reflected evidence. In contrast, \nickname based on SEVA leverages reflected views to generate content revealed in the mirror.

\begin{table}[t!]
\centering
\footnotesize
\caption{PSNR comparison on the six scenes of synthetic data with sparse-input setting. Our method shows larger performance gap in the reflected region than in the full scene.}
\renewcommand{\arraystretch}{0.6}
\setlength{\tabcolsep}{3pt}
\begin{tabular}{lccc}
\toprule
Method & Full & Reflected region & Mirror surface\\
\midrule
MVGenMaster & 16.90 & 16.51 & 19.41 \\
Ours (MVGen.) & 17.95 {\textcolor{blue!60}{(+1.05)}} & 18.40 {\textcolor{blue!60}{(+1.89)}} & 19.69 {\textcolor{blue!60}{(+0.28)}} \\
\midrule
SEVA & 13.65 & 13.31 & 15.25 \\
Ours (SEVA) & 15.51 {\textcolor{blue!60}{(+1.86)}} & 16.11 {\textcolor{blue!60}{(+2.80)}} & 16.16 {\textcolor{blue!60}{(+0.91)}} \\

\bottomrule
\end{tabular}
\label{tab:region_aware}
\end{table}
\begin{figure}[t!]
    \centering
    \includegraphics[width=1\linewidth]{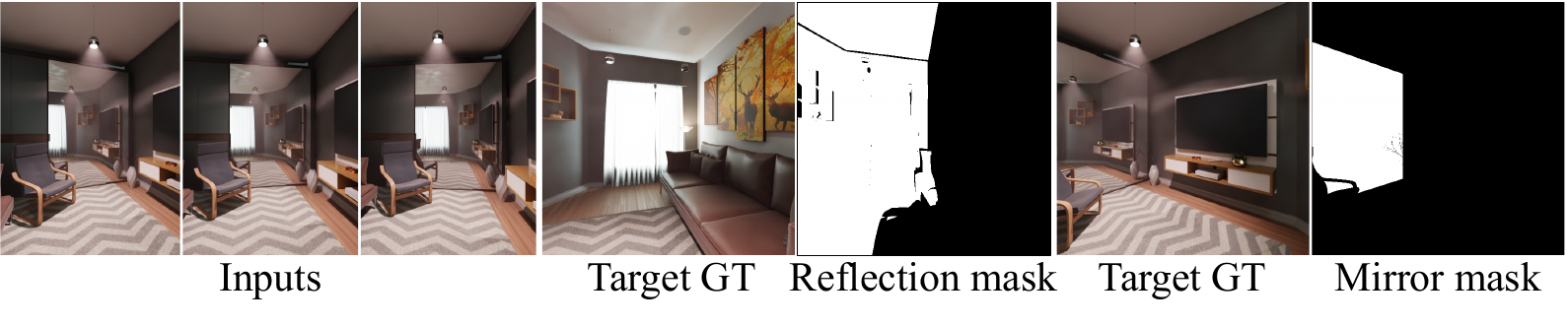}
    \caption{Examples of reflection mask and mirror mask. We extract ground truth binary masks of reflected region of the scene and mirror region using Blender~\cite{blender}.}
    \label{fig:rebut_region_aware}
\end{figure}
\paragraph{Region-wise evaluation}
We measure region-wise PSNR by masking the reflected regions and the mirror surface (shown in Fig.~\ref{fig:rebut_region_aware}). Table~\ref{tab:region_aware} shows that the performance gap is greater in the reflected region than in the full scene, and the mirror surface generation also improves.
Note that the PSNR in the reflected region can remain low due to large camera pose gaps between target and reflected input views, making it a wide-baseline NVS setting.

\paragraph{Inference time}
For the sparse-image novel view synthesis setting, \nickname based on SEVA requires approximately 397 seconds per scene, while SEVA requires about 130 seconds in wall-clock time. 
The additional computation arises from the automated preprocess and two-step generation pipeline, and can be viewed as a computational trade-off for enabling reflection-consistent view synthesis without additional training.
All runtimes are measured on an NVIDIA RTX 6000 Ada Generation GPU.


%
%
\bibliographystyle{splncs04}
\bibliography{references}